\PassOptionsToPackage{table}{xcolor}
\documentclass{article}
\usepackage{preprint,times}

\usepackage{amsmath,amsfonts,bm}

\def\eqref#1{equation~\ref{#1}}

\def\1{\bm{1}}

\DeclareMathAlphabet{\mathsfit}{\encodingdefault}{\sfdefault}{m}{sl}
\SetMathAlphabet{\mathsfit}{bold}{\encodingdefault}{\sfdefault}{bx}{n}

\usepackage{booktabs}
\usepackage{graphicx}
\usepackage{float}
\usepackage{wrapfig}
\usepackage{tikz}
\usepackage{pgfplots}\pgfplotsset{compat=1.18}
\definecolor{bbNavy}{HTML}{2E4A7D}\definecolor{bbGray}{HTML}{A3A3A3}
\definecolor{famG}{HTML}{7F96C2}\definecolor{famC}{HTML}{C98A84}\definecolor{famM}{HTML}{D6BF7A}
\graphicspath{{resources/figures/}}
\usepackage{multirow}
\usepackage{subcaption}
\usepackage[table]{xcolor}
\usepackage[normalem]{ulem}
\usepackage{enumitem}
\usepackage{hyperref}
\usepackage[capitalize,noabbrev]{cleveref}
\usepackage{titletoc}
\titlecontents{section}[1.8em]{\addvspace{7pt}\bfseries\color{cite_blue}}{\contentslabel{1.8em}}{}{\titlerule*[0.6pc]{.}\contentspage}
\titlecontents{subsection}[4.4em]{\addvspace{1pt}\color{cite_blue}}{\contentslabel{2.6em}}{}{\titlerule*[0.6pc]{.}\contentspage}
\hypersetup{linktoc=all}
\usepackage{url}
\usepackage{xurl}

\newcommand{\mname}{\text{ATLAS}}
\newcommand{\benchname}{\text{MAABench}}

\newcommand{\best}[1]{\textbf{#1}}
\newcommand{\second}[1]{\uline{#1}}

\newcommand{\huggingface}{\raisebox{-1.5pt}{\includegraphics[height=1.05em]{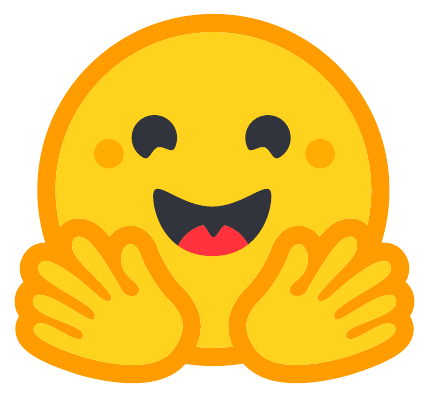}}}
\newcommand{\github}{\raisebox{-1.5pt}{\includegraphics[height=1.05em]{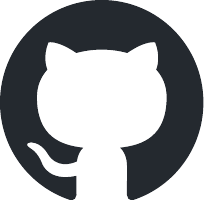}}}
\newcommand{\worldwideweb}{\raisebox{-1.5pt}{\includegraphics[height=1.05em]{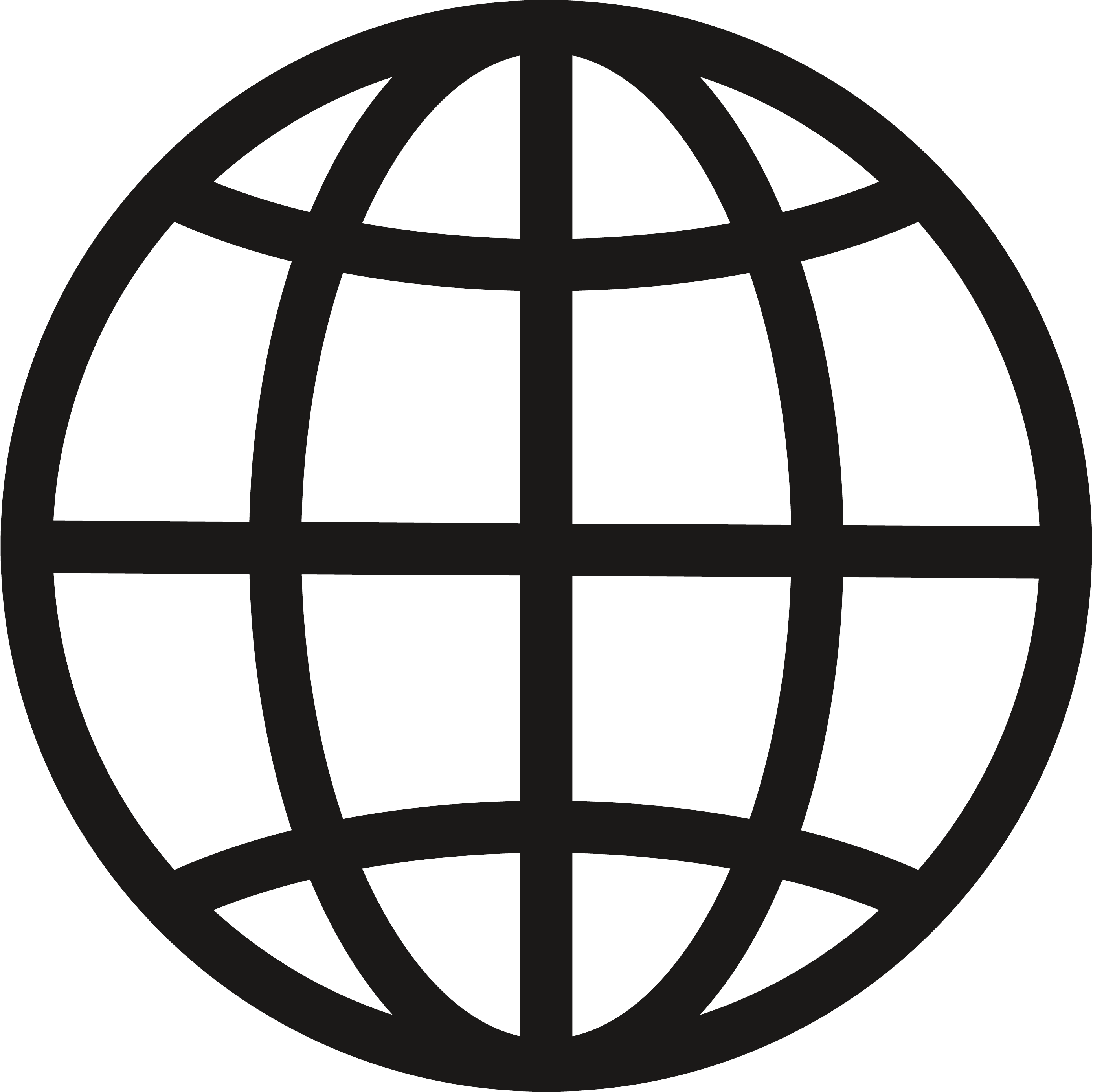}} }

\title{Who's Behind the Text? \\Robust Model Authorship Attribution}

\shorttitle{Who's Behind the Text? Robust Model Authorship Attribution}

\author{
Minseok~Ko\textsuperscript{*}\ \ 
Hyunseok~Lee\textsuperscript{*}\ \
Woomin~Song\ \ \
Beomjun~Kim\ \ \
Jinwoo~Shin\ \ \
Kunwoong~Kim
}
\affiliation{Korea Advanced Institute of Science and Technology (KAIST)}
\correspondence{\textsuperscript{*}Equal contribution. Correspondence to: \href{mailto:koms0109@kaist.ac.kr,kunwoong.kim@kaist.ac.kr}{\{koms0109, kunwoong.kim\}@kaist.ac.kr}.}
\logo{kaist_ai_logo.pdf}

\hypersetup{
  colorlinks = true,
  citecolor = {cite_blue},
  linkcolor = {cite_blue},
  urlcolor = {cite_blue},
}

\begin{document}
\maketitle

\begin{abstract}
    As large language models (LLMs) are increasingly used to generate text, their misuse has become a growing societal problem, with implications for academic misconduct, model distillation, and cyberattacks.
    For these reasons, Model Authorship Attribution (MAA), which identifies the source model of a given text, is important for promoting the responsible use of LLMs and improving AI safety.
    In practice, attribution systems must remain reliable beyond the distributions observed during training. They may encounter texts from unseen domains and writing settings, as well as generations from future versions of model families that were not available when the system was trained.
    However, existing MAA methods often exhibit substantial performance degradation under such distribution shifts.
    To address this challenge, we propose \textbf{A}uthorship \textbf{T}racing via \textbf{L}ikelihood-based \textbf{A}ttribution of \textbf{S}ources (\mname{}),
    which formulates MAA through source-conditioned text modeling rather than direct source prediction. \mname{} learns how text from each source family is generated and attributes a given text by comparing its likelihood under each learned source condition. We aggregate this likelihood evidence at the token level with clipping so that no single token dominates the decision.
    We further introduce \benchname{}, a broad MAA benchmark for evaluating practical OOD settings such as domain and version.
    Our experiments show that \mname{} consistently outperforms existing baselines across \benchname{} and OpenTuringBench.
    In particular, when both the generation domain and the model version change on \benchname{}, \mname{} achieves 74.27\% accuracy, 9.00pp higher than the strongest baseline.
    Beyond MAA, \mname{} can also help identify distillation source models by measuring which source model a target LLM most closely resembles.
    Our extended experiments show that \mname{} recovers the known teacher of every distilled model we evaluate.
\end{abstract}

\begin{center}
    \renewcommand{\arraystretch}{1.5}
    \begin{tabular}{rll}
        \worldwideweb{} & \textbf{Website} & \href{https://komseok0109.github.io/ATLAS/}{\nolinkurl{komseok0109.github.io/ATLAS/}}\\
        \huggingface{} & \textbf{Tutorial page} & \href{https://huggingface.co/spaces/ming9999/ATLAS-demo}{\nolinkurl{huggingface.co/spaces/ming9999/ATLAS-demo}} \\
        \github{} & \textbf{Experiment code} & \href{https://github.com/komseok0109/ATLAS}{\nolinkurl{github.com/komseok0109/ATLAS}}\\
    \end{tabular}
\end{center}

\section{Introduction}
\label{sec:intro}

Large language models (LLMs) are now widely used across a range of writing tasks, including news, reviews, essays, and technical writing~\citep{survey}.
Accordingly, considerable attention has been devoted to machine-generated text (MGT) detection, which distinguishes LLM-generated text from human-written text~\citep{detectgpt,fastdetectgpt,binoculars,remodetect}.
However, when harmful, misleading, or otherwise problematic content is generated by an LLM, merely identifying that the text is machine-generated is often insufficient.
Further investigation may require identifying \emph{which model} generated the text, for example, to trace the model or provider associated with the generated content.

This motivates the more fine-grained problem of \textit{Model Authorship Attribution (MAA)}, which aims to identify the source model family of a given text.
Beyond tracing the source of individual texts, MAA may also provide information about relationships between models.
A particularly interesting example is output-based distillation, where a student model is trained on outputs generated by a teacher model.
If source-specific characteristics of the teacher remain alive in the student's generations, attribution signals may provide clues about the teacher model.
Recent incidents involving the reuse of frontier-model outputs further motivate the use of source attribution~\citep{claude_misuse}.

A key practical challenge, however, is whether attribution remains reliable as the conditions under which text is generated change over time.
A deployed attribution system may encounter texts from domains and writing settings that were absent during training, making it important to distinguish persistent source-specific signals from domain- or task-specific cues.
Meanwhile, model families evolve rapidly, and new versions are continually released day by day.
Retraining an attribution system whenever a source model is updated can be costly and impractical.
Ideally, a system trained today should remain able to attribute texts produced by future versions of the same model family that were not available during training.

Existing MAA methods~\citep{sniffer,gpt-who,detective,openturingbench,idiosyncrasies} perform well when the training and test distributions are similar, but often degrades under such changes.
Because they directly predict source model from text, they may rely on source-discriminative cues that are unstable under distribution shift.
Motivated by prior evidence that generative modeling can be more robust to such shortcut solutions~\citep{generative}, we instead approach MAA from the generative and likelihood-based perspective.

\begin{figure}[t]
\centering
\includegraphics[width=0.9\textwidth]{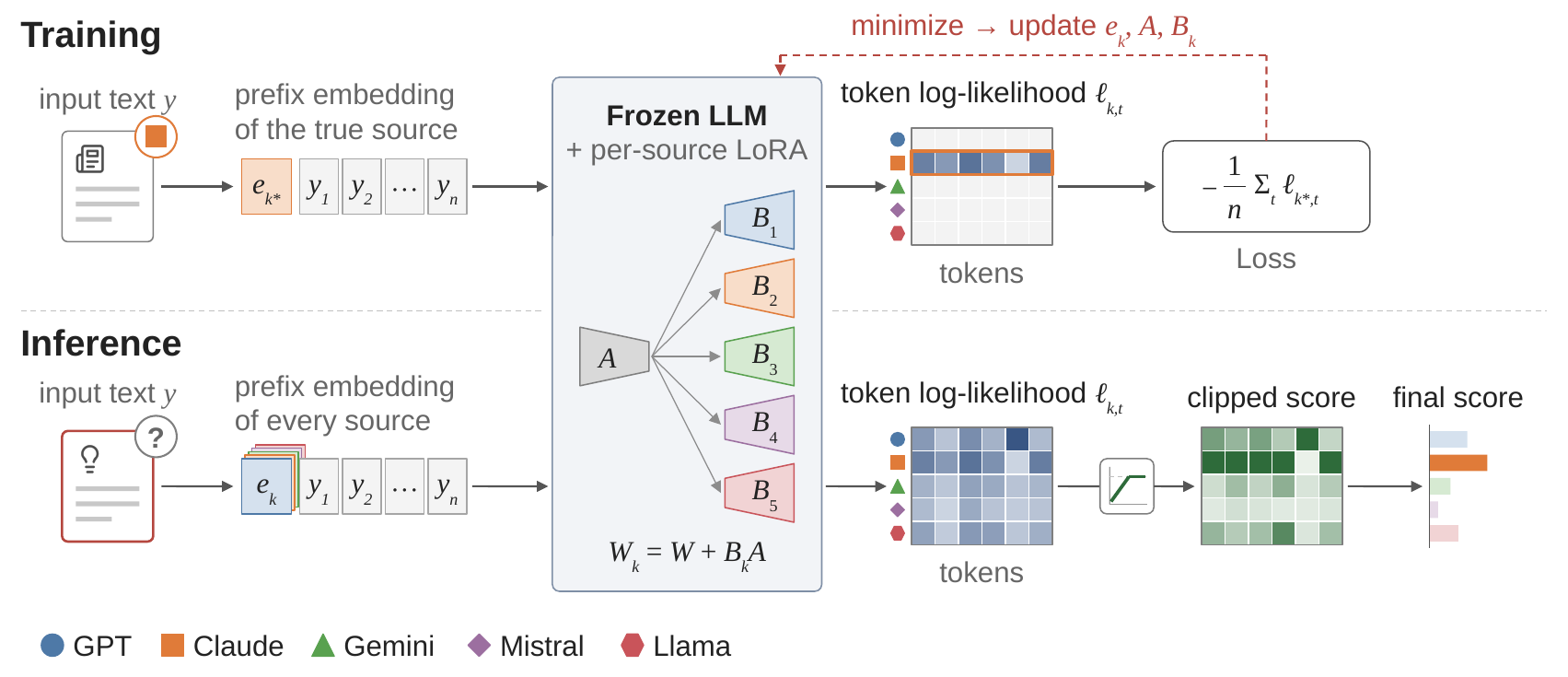}
\caption{
\textbf{Overview of \mname{}.}
Rather than directly predicting the source, \mname{} learns a source-conditioned language model for each model family and performs attribution from their relative token likelihoods.
During training, for each source $k$, \mname{} adapts a pretrained LLM with a prefix embedding $e_k$ and a LoRA update $W_k = W + B_kA$, where $A$ is shared down-projection and $B_k$ is source-specific up-projection.
For the ground-truth source $k^*$, \mname{} optimizes $e_{k^*}$, $B_{k^*}$, and the shared $A$ using the mean negative log-likelihood.
During inference, \mname{} calculates token log-likelihoods for all source $k$, which serve as source model scores after clipping.
}
\label{fig:overview}
\vskip -0.05in
\end{figure}

\paragraph{Our proposed framework and benchmark.}

In this work, we propose \textbf{A}uthorship \textbf{T}racing via
\textbf{L}ikelihood-based \textbf{A}ttribution of \textbf{S}ources
(\mname{}), a likelihood-based MAA framework based on lightweight adaptations of a pretrained LLM.
Rather than directly predicting the $k$-th source family
from a given text $y$, \mname{} learns a source-conditioned model of the text distribution for each family and performs attribution by comparing their likelihoods.
Furthermore, to prevent a small number of tokens with extreme likelihood differences from dominating the attribution, \mname{} compares source likelihoods at the token level and clips their relative evidence.
Figure~\ref{fig:overview} illustrates \mname{}.

We also introduce \benchname{}, a new benchmark for evaluating MAA beyond the training distribution.
Existing MAA benchmarks cover only part of this setting~\citep{turingbench,openturingbench,multilingual}, while other benchmarks that explicitly consider model version shift only focus on MGT detection~\citep{evobench}.
\benchname{} evaluates two major types of distribution shift (i.e., generation domain and source-model version).
It contains texts generated by five recent LLM families, GPT~\citep{gpt5}, Claude~\citep{claude}, Gemini~\citep{gemini}, Mistral~\citep{mistral}, and Llama~\citep{llama}, across six domains.

Across \benchname{} and OpenTuringBench~\citep{openturingbench}, \mname{} consistently outperforms existing baselines under various Out-Of-Distribution (OOD) settings, as expected in our motivation for source-conditioned likelihood modeling.
In the most challenging setting of \benchname{}, where both the generation domain and source-model version change, \mname{} improves over the strongest baseline by 9.00pp.

Interestingly, we further show that the source model scores learned by \mname{} can be useful beyond direct MAA.
In specific, we use them to identify the teacher of a distilled model of interest, asking whether the outputs of a student model are similar to its teacher.
On publicly available student models distilled from known teachers, \mname{} correctly identifies the teacher of every model, even when students of the same teacher are built on different base models.

Our main contributions are summarized as follows:
\begin{itemize}[nosep,leftmargin=0.2in,itemsep=0.15in]
    \item
        \textbf{A likelihood-based method of model authorship attribution (MAA).}
        We propose \mname{}, which learns source-conditioned text distributions with lightweight adaptations and performs attribution from the token-level likelihoods.
        We further introduce clipping the token-level likelihoods for robust aggregation resulting in the final source model scores.

    \item
        \textbf{A new MAA benchmark covering various distribution shifts.}
        We introduce \benchname{}, which evaluates distribution shifts in generation domain, model version, and their combination.

    \item
        \textbf{Strong empirical performance under the distribution shifts.}
        Experiments on \benchname{} and OpenTuringBench~\citep{openturingbench} show that \mname{} consistently improves OOD attribution over existing baselines.

    \item
        \textbf{Source attribution for various distilled and deployed models.}
        We show that the source signals learned by \mname{} retain teacher-specific information after distillation, enabling teacher identification for distilled models.
\end{itemize}

\section{Related Work}
\label{sec:related}

\paragraph{Machine-generated text (MGT) detection.}
One line of MGT detection trains supervised binary classifiers to distinguish MGT from human-written text~\citep{gptsentinel}.
However, they are often fragile under out-of-distribution (OOD) settings, limiting their practical applicability~\citep{detective}.
To address these limitations, diverse approaches have been proposed:
(i) Exploit signals from pretrained language models: probability curvature~\citep{detectgpt,fastdetectgpt}, n-gram divergence~\citep{dnagpt}, paired-model likelihoods~\citep{binoculars}, and log-rank~\citep{detectllm} statistics.
These signals are often computed under a surrogate model whose mismatch with the generator can motivate routing to a better-matched model~\citep{minimizing}.
(ii) Watermarking~\citep{watermark} add perturbation during generation, but applies only when a provider chooses to embed a mark.
(iii) Other approaches construct learned detection signals, for example using reward-based scores~\citep{remodetect,irm} or representations of writing style~\citep{detective}.
In contrast, \mname{} derives attribution signal from source-conditioned token-level likelihoods.
Moreover, while these works ask whether a given text is machine-generated, we ask which model wrote it.

\paragraph{Model authorship attribution (MAA).}
Beyond binary MGT detection, recent work studies the finer-grained task of identifying the source model of generated text.
One line relies on statistics computed under frozen models: Sniffer~\citep{sniffer} contrasts model-specific features, GPT-who~\citep{gpt-who} uses uniform information density, and LLMDet~\citep{llmdet} estimates source-specific likelihoods from token statistics.
Another learns text representations directly, using contrastive objectives to cluster texts by source~\citep{detective,openturingbench}, fine-tune text embedding models~\citep{idiosyncrasies}, formulate attribution as next-token prediction~\citep{t5sentinel}, jointly model detection and attribution~\citep{damtl}, or adapt to new sources without full retraining~\citep{ridgeft}.
Existing MAA methods largely rely on source-discriminative signals, which can be unstable under distribution shift. In contrast, \mname{} learns source-conditioned text distributions and is more robust to domain and version shifts.

\paragraph{Benchmarks under distribution shift.}
Benchmarks for MGT detection have expanded from unified evaluations~\citep{mgtbench} to broader coverage of generators and domains~\citep{m4,detectanyllm}, adversarial and real-world variation~\citep{raid,detectrl}, and evolving model versions~\citep{evobench}.
MAA benchmarks have similarly moved from attribution among a fixed set of generators~\citep{turingbench} to in-domain variations and a domain shift~\citep{openturingbench} and to multilingual attribution~\citep{multilingual}.
However, none of the MAA benchmarks considers version changes within the same model family, and EvoBench \citep{evobench}, which does, targets MGT detection rather than MAA.
\benchname{} evaluates domain and version changes, separately and together, within one benchmark for MAA.

\clearpage
\section{A Likelihood-Based Framework for Authorship Attribution}
\label{sec:method}

\subsection{Problem Setup: Model Authorship Attribution}
\label{sec:problem}

The goal of Model Authorship Attribution (MAA) is to identify which language model family generated a given text.
Let $\mathcal{M}_{1}, \ldots, \mathcal{M}_{K}$ be the $K$ candidate language model families (e.g., GPT, Claude, and Gemini), where $\mathcal{M}_k$ for $k\in\{1,\ldots,K\}$ denotes the set of models belonging to family $k$.
For example, if family $k$ corresponds to GPT, then $\mathcal{M}_k=\{\ldots,\textup{GPT-3},\ldots,\textup{GPT-5.6},\textup{GPT-6},\ldots\}$.

Given a text $y\in\mathcal{Y}$, we assume that $y$ is generated from $p_M(\cdot\mid g)$ for some source model $M\in\mathcal{M}:=\bigcup_{k=1}^{K}\mathcal{M}_k$ and generation condition $g\in\mathcal{G}$.
Here, $g$ captures factors other than the source-model identity that may affect generation, such as the domain, prompt, and available context.
We say that the source family of $y$ is $k$ if $M\in\mathcal{M}_k$.
The objective of MAA is therefore to learn an attribution function $f:\mathcal{Y}\to\{1,\ldots,K\}$ that predicts the source family of a given text as $\hat{k}:=f(y)$.

Not only the In-Domain (ID) setting (train distribution of $y$ $\approx$ test distribution of $y$), but we consider Out-Of-Distribution (OOD) settings, where texts given at inference time differ from the training texts, in terms of their generation conditions, source-model versions, or both.
This includes cases involving previously unseen generation conditions or new versions of known source-model families.

\subsection{\mname{}}

We propose a new approach to model authorship attribution, called \mname{} (Authorship Tracing via Likelihood-based Attribution of Sources).
A key idea of \mname{} is to formulate MAA through source-conditioned text modeling rather than direct source prediction. For each source family \(\mathcal{M}_k\), \mname{} learns a source condition \(c_k\) that models the likelihood of texts generated by that family. Given a text \(y\), attribution is performed by comparing how well the different source conditions explain the observed text.
Prior work has shown that such class-conditional generative modeling can be less reliant on shortcut features under distribution shift than direct discriminative prediction~\citep{generative}.
In particular, we compare source conditions individually (i.e., token by token), rather than aggregating token likelihoods.
Then, the log-likelihood of each token is clipped, so a few tokens with extreme log-likelihoods cannot determine the attribution solely.

\paragraph{Training: source-specific adaptation.}

During training, for a text $y=(y_1,\ldots,y_n)$ with $n$ tokens and its ground-truth source family $k^*=k^*(y)$, we maximize its likelihood under the corresponding source condition.
The same source condition $c_k$ is used for all training texts from family $k$, irrespective of the particular generation settings represented in the training data.
That is, we consider the mean negative log-likelihood as the training objective:
$\mathcal{L}(y)=-\frac{1}{n}\sum_{t=1}^{n}\ell_{k^*,t}$,
where
$\ell_{k,t}=\log p (y_t\mid y_{<t},c_k)$
denotes the $t$-th token log-likelihood under source condition $c_k$.
Importantly, \mname{} does not use a loss that directly predicts the source label.  The source label is used only to select the condition under which the text likelihood is optimized.

We implement the source conditions within a frozen pretrained backbone using a learnable prefix embedding and source-specific LoRA adaptation~\citep{lora}.
That is, for each source $k$, we train $c_k=(e_k,B_k)$, where $e_k$ is a learnable source-specific prefix embedding and $B_k$ is the source-specific LoRA up-projection.
For each backbone weight $W$, the source-conditioned weight is then computed as $W_k=W+\Delta W_k$, where $\Delta W_k=B_kA$ and $A$ is the down-projection shared across all sources.
We note that this conditioning technique is also related to the asymmetric and efficient LoRA adaptation~\citep{hydralora,asymmetry}.

\paragraph{Inference:
aggregating source-conditioned likelihood evidence.
}

At inference time, the given text is evaluated under all $K$ source conditions.
A natural approach is to average the token log-likelihoods $\ell_{k,t}=\log p (y_t\mid y_{<t},c_k)$ for all $t = 1, \ldots, n$ and select the source family with the highest mean likelihood.
However, this makes the attribution score sensitive to a small number of tokens with extreme likelihood differences. We therefore compare the source conditions token by token and bound the contribution of each token through clipped relative likelihoods.

For each source $k$ and token $y_t$, we compare its likelihood against every competing source and clip each pairwise ratio at one.
This yields a bounded score $r_{k,t}\in(0,1]$ that equals $1$ when source $k$ assigns the highest likelihood and decreases as competing sources explain the token better, preventing a few tokens from dominating the final attribution score.
Because $r_{k,t}$ depends only on relative log-likelihood differences, it is invariant to source-independent additive shifts.
We average these scores across tokens and temperatures to obtain $R_k$, and predict $\hat{k}=\arg\max_k R_k$.
The exact formulations are given in Appendix~\ref{app:method}.

\section{\benchname}
\label{sec:bench}

This section introduces \benchname{}, a new benchmark for evaluating MAA under distribution shift that covers a practically important setting that previous benchmarks did not.
We consider two types of distribution shift: the generation domain and the source model version.
Based on these two types, we build three OOD settings where the domain changes, the model version changes, or both change.

\subsection{Data Construction}
\label{sec:benchmark_data}

\paragraph{Model families and generation domains.}

We construct \benchname{} using generations from five recent model families:
GPT~\citep{gpt5}, Claude~\citep{claude}, Gemini~\citep{gemini}, Mistral~\citep{mistral}, and Llama~\citep{llama}.
The dataset contains 4 model versions for each model family, corresponding to different release versions.
Specifically, in the version-OOD setting, the oldest version is used for training, and the remaining three versions are used for test.

For the generation domains, we consider six:
News~\citep{xsum}, Wikipedia~\citep{wikidump}, Academic Writing~\citep{arxiv}, Product Review~\citep{review}, Patent~\citep{hupd}, and Stack Exchange~\citep{stackexchange}.
In detail, for the domain-OOD setting, we use News, Wikipedia, Academic Writing, and Product Review for the training data, and the remaining two domains (i.e., Patent and Stack Exchange) are used for test.

\paragraph{Construction details.}

For each domain, we construct generation prompts from the metadata that accompanies each human-written text in its source corpus (e.g., the headline of a news article) and give the same prompt to all five families, yielding prompt-paired generations; the human-written texts are kept for MGT detection.
Moreover, following \citet{detectanyllm}, we apply the same length constraint and filtering procedure to all source models, targeting 90-220 words and retaining only complete prompt-paired groups that pass the filter.

\subsection{Evaluation Settings}
\label{sec:benchmark_settings}

\begin{wraptable}{r}{0.48\textwidth}
\vspace{-0.13in}
\centering
\caption{
\textbf{Summary of \benchname{}.}
}
\label{tab:benchmark_overview}
\footnotesize
\resizebox{0.95\linewidth}{!}{%
\begin{tabular}{lcccc}
\toprule
\textbf{Distribution}
& \textbf{Split}
& \textbf{Setting}
& \textbf{MGT}
& \textbf{Human} \\
\midrule

\multirow{3}{*}{ID}
& Train
& --
& 149,760 & 29,952 \\
& Val
& --
& 9,990 & 1,998 \\
& Test
& --
& 9,985 & 1,997 \\

\midrule

\multirow{3}{*}{OOD}
& \multirow{3}{*}{Test}
& Domain
& 7,470 & 1,494 \\
&
& Version
& 29,700 & 1,980$^{\dagger}$ \\
&
& Joint
& 21,615 & 1,441$^{\dagger}$ \\

\midrule
\multicolumn{3}{l}{\textbf{Total}}
& \textbf{228,520}
& \textbf{35,443} \\
\bottomrule
\end{tabular}
}
\vspace{-0.1in}
\end{wraptable}

We evaluate one in-distribution (ID) setting and three out-of-distribution (OOD) settings.
\emph{ID} uses the domains and source-model versions observed during training.
\emph{Domain shift} uses the held-out domains, generated with a persona-based system prompt instead of the default one (Appendix~\ref{sec:bench_prompt}), while keeping the training-time source-model versions fixed.
\emph{Version shift} uses the training domains but replaces each training-time source model with one of three held-out versions from the same family.
\emph{Joint shift} are the shift when both the Domain shift and Version shift jointly occur at the same time.
Table~\ref{tab:benchmark_overview} summarizes the resulting dataset with corresponding splits and evaluation settings, which contains 228,520 machine-generated texts and 35,443 human-written texts.
Further details on \benchname{} are provided in Appendix~\ref{app:bench}.

\section{Experiments}
\label{sec:experiments}

In this section, we empirically evaluate \mname{} from three perspectives.
First, we compare \mname{} with existing MAA methods on \benchname{} and OpenTuringBench under both ID and OOD settings.
Second, we examine whether the source-conditioned likelihoods learned by \mname{} can serve as good signals for binary MGT detection.
Finally, we analyze the contribution of each component of \mname{} and how its performance depends on the backbone model and the training data size.
Further analyses of these results are provided in Appendix~\ref{app:exp_additional}.

\subsection{Experimental Setup}
\label{sec:exp_setup}

\paragraph{Implementation details and evaluation setup.}

For the main experiments, we use Qwen2.5-7B~\citep{qwen2.5} as the backbone for \mname{}.
On \benchname{}, the models of all methods are trained on the ID training data and evaluated on the ID and three OOD test data (Domain, Version, and Joint).
On OpenTuringBench~\citep{openturingbench}, we follow its evaluation setting (i.e., on the original ID, four ID variations (E1-E4), and OOD).
We report accuracy and macro-F1 in main MAA experiments.
For the MGT detection, we report AUROC, the true positive rate at a $1\%$ false positive rate (TPR@1\%FPR), and the false positive rate at a $95\%$ true positive rate (FPR@95\%TPR), with MGT as the positive class.
Further implementation details are in Appendix~\ref{app:exp_training}.

\paragraph{Baselines.}

We compare \mname{} with nine MAA baselines.
Five methods use statistics from proxy models:
Fast-DetectGPT~\citep{fastdetectgpt},
Binoculars~\citep{binoculars},
GPT-who~\citep{gpt-who},
Sniffer~\citep{sniffer},
and StatEnsemble~\citep{multilingual}.
The remaining four methods learn representations for attribution:
DeTeCtive~\citep{detective},
OTBDetector~\citep{openturingbench},
DA-MTL~\citep{damtl},
and Idiosyncrasies~\citep{idiosyncrasies}.
For MGT detection, we additionally compare with ReMoDetect~\citep{remodetect} and IRM~\citep{irm}, which detect MGT by using reward signals.
Detailed implementations are in Appendix~\ref{app:exp_baselines}.

\begin{table*}[t]
\centering
\footnotesize
\setlength{\tabcolsep}{4.2pt}
\renewcommand{\arraystretch}{1.05}

\caption{
\textbf{MAA performance on \benchname{}.}
We report accuracy (\%, $\uparrow$) and macro-F1 (\%, $\uparrow$).
Best results are in \textbf{bold}.
For \mname{} and the four encoder-tuning baselines, we report the mean over three random seeds, with standard deviations and per-seed results in Table~\ref{tab:main_seeds}; the remaining baselines are from a single run.
}
\label{tab:main_results}

\begin{tabular}{
@{}l
@{\hspace{8pt}}
cc
@{\hspace{5pt}}
cc
@{\hspace{5pt}}
cc
@{\hspace{5pt}}
cc@{}
}
\toprule

\multirow{3}{*}{Method}
& \multicolumn{2}{c}{\multirow{2}{*}{ID}}
& \multicolumn{6}{c}{OOD} \\
\cmidrule(lr){4-9}

&
&
& \multicolumn{2}{c}{Domain}
& \multicolumn{2}{c}{Version}
& \multicolumn{2}{c}{Joint} \\

\cmidrule(lr){2-3}
\cmidrule(lr){4-5}
\cmidrule(lr){6-7}
\cmidrule(lr){8-9}

& Acc & F1
& Acc & F1
& Acc & F1
& Acc & F1 \\

\midrule

Fast-DetectGPT~\citep{fastdetectgpt}
& 44.37 & 43.54
& 45.29 & 44.18
& 35.22 & 34.75
& 31.85 & 30.81 \\

Binoculars~\citep{binoculars}
& 45.61 & 44.81
& 46.53 & 45.70
& 36.09 & 35.30
& 32.56 & 31.17 \\

GPT-who~\citep{gpt-who}
& 58.63 & 58.29
& 55.29 & 56.17
& 41.71 & 40.34
& 39.94 & 38.43 \\

Sniffer~\citep{sniffer}
& 76.00 & 75.83
& 64.51 & 64.89
& 51.97 & 50.72
& 47.57 & 45.85 \\

StatEnsemble~\citep{multilingual}
& 74.36 & 74.11
& 58.07 & 58.52
& 38.86 & 38.77
& 32.29 & 32.35 \\

\addlinespace[2pt]

DeTeCtive~\citep{detective}
& 99.51 & 99.51
& 94.69 & 94.72
& 70.60 & 70.21
& 63.32 & 62.63 \\

OTBDetector~\citep{openturingbench}
& 99.50 & 99.50
& 94.89 & 94.89
& 70.20 & 69.73
& 63.48 & 62.54 \\

DA-MTL~\citep{damtl}
& 99.58 & 99.58
& 95.31 & 95.31
& 71.08 & 70.78
& 64.99 & 64.02 \\

Idiosyncrasies~\citep{idiosyncrasies}
& 99.56 & 99.56
& 95.04 & 95.05
& 71.98 & 71.61
& 65.27 & 64.11 \\

\midrule
\rowcolor{gray!10}
\mname{} (ours)
& \best{99.72} & \best{99.72}
& \best{97.97} & \best{97.97}
& \best{84.30} & \best{84.27}
& \best{74.27} & \best{73.99} \\

\bottomrule
\end{tabular}
\end{table*}

\subsection{Main Results}
\label{sec:main_results}

\paragraph{MAA performance}

Table~\ref{tab:main_results} compares \mname{} with the nine baselines on \benchname{}.
While the ID setting is nearly saturated (i.e., \mname{} and some of baselines achieve over 99\% accuracy), the performance gaps are substantial under OOD evaluation.
That is, \mname{} outperforms the strongest baseline by 2.66pp, 12.32pp, and 9.00pp on Domain shift, Version shift, and Joint shift, respectively.
The two OOD types also affect \mname{} differently: relative to ID, accuracy decreases by only 1.75pp on Domain shift, but with 15.42pp on Version shift.
These results imply that source attribution can generalize more readily across the generation domains than across unseen versions.

Table~\ref{tab:otb_results} further evaluates the same methods on OpenTuringBench~\citep{openturingbench}.
\mname{} achieves the strongest OOD performance, achieving 95.71\% accuracy and outperforming the best baseline by 33.52pp.
It also achieves the best performance on all four ID variations (i.e., E1-E4), showing that the improvement of \mname{} is not tied to our benchmark \benchname{}.
Notably, the advantage of \mname{} becomes substantially larger in the OOD setup, aligned with our motivation for modeling source-conditioned likelihoods instead of directly optimizing source attribution.

\begin{table*}[t]
\centering
\footnotesize
\setlength{\tabcolsep}{4.2pt}
\renewcommand{\arraystretch}{1.05}

\caption{
\textbf{MAA performance on OpenTuringBench.}
We report accuracy (\%, $\uparrow$), and E1-E4 are the in-domain variation tasks.
Best results are in \textbf{bold}.
}
\label{tab:otb_results}

\begin{tabular}{
@{}l
@{\hspace{8pt}}
ccccc
@{\hspace{8pt}}
c@{}
}
\toprule

\multirow{2}{*}{Method}
& \multicolumn{5}{c}{ID}
& \multirow{2}{*}{OOD} \\

\cmidrule(lr){2-6}

& ID & E1 & E2 & E3 & E4 & \\

\midrule

Fast-DetectGPT~\citep{fastdetectgpt}
& 18.93 & 19.49 & 17.57 & 21.68 & 15.37
& 19.92 \\

Binoculars~\citep{binoculars}
& 19.54 & 22.91 & 20.33 & 24.79 & 16.41
& 23.15 \\

GPT-who~\citep{gpt-who}
& 47.55 & 22.41 & 44.68 & 31.09 & 15.05
& 25.89 \\

Sniffer~\citep{sniffer}
& 61.65 & 27.25 & 55.78 & 38.67 & 15.53
& 30.12 \\

StatEnsemble~\citep{multilingual}
& 61.93 & 60.52 & 58.40 & 63.30 & 26.99
& 48.46 \\

\addlinespace[2pt]

DeTeCtive~\citep{detective}
& {99.83} & {84.43} & 99.36 & 89.00 & 27.48
& 56.61 \\

OTBDetector~\citep{openturingbench}
& 99.76 & 84.41 & 99.11 & 88.03 & 30.89
& 47.92 \\

DA-MTL~\citep{damtl}
& \best{99.87} & 83.79 & {99.48} & {95.99} & {32.36}
& {62.19} \\

Idiosyncrasies~\citep{idiosyncrasies}
& 89.90 & 70.16 & 86.56 & 79.52 & 28.64
& 41.11 \\

\midrule
\rowcolor{gray!10}
\mname{} (ours)
& 99.80 & \best{89.71} & \best{99.70} & \best{98.79} & \best{41.92}
& \best{95.71} \\

\bottomrule
\end{tabular}
\end{table*}

\paragraph{MGT detection performance}

As a secondary analysis, we examine whether the source-conditioned likelihoods learned by \mname{} are useful for MGT detection.
Table~\ref{tab:mgt_results} shows that, \mname{} performs competitive to the strongest detectors trained on \benchname{} across all cases, suggesting that one can apply \mname{} not only to MAA but also to the MGT detection.
Table~\ref{tab:mgt_tpr} in Appendix~\ref{app:exp_mgt} shows the similar results, with respect to TPR@1\%FPR and FPR@95\%TPR.

\begin{table}[t]
\centering
\footnotesize
\setlength{\tabcolsep}{4.2pt}
\renewcommand{\arraystretch}{1.05}
\caption{
\textbf{MGT detection on \benchname{}.}
We report AUROC (\%, $\uparrow$) with MGT as the positive class; \emph{Trained on \benchname{}} marks methods trained on the \benchname{} training split, while the others use off-the-shelf models.
Best results are in \textbf{bold}.
}
\label{tab:mgt_results}

\begin{tabular}{
@{}l
@{\hspace{8pt}}
c
@{\hspace{8pt}}
c
@{\hspace{5pt}}
ccc@{}
}
\toprule

\multirow{2}{*}{Method}
& \multirow{2}{*}{\shortstack{Trained on\\\benchname{}}}
& \multirow{2}{*}{ID}
& \multicolumn{3}{c}{OOD} \\

\cmidrule(lr){4-6}

& & & Domain & Version & Joint \\

\midrule

Fast-DetectGPT~\citep{fastdetectgpt}
& $\times$
& 65.59
& 65.25 & 57.56 & 50.74 \\

Binoculars~\citep{binoculars}
& $\times$
& 71.14
& 71.74 & 62.56 & 57.57 \\

IRM~\citep{irm}
& $\times$
& 73.85
& 62.62 & 70.16 & 54.31 \\

ReMoDetect~\citep{remodetect}
& $\times$
& 77.27
& 82.33 & 73.06 & 80.32 \\

\addlinespace[2pt]

DeTeCtive~\citep{detective}
& \checkmark
& \best{100.00}
& 98.60 & 99.95 & 97.97 \\

OTBDetector~\citep{openturingbench}
& \checkmark
& 99.94
& 99.39 & 99.80 & 96.81 \\

DA-MTL~\citep{damtl}
& \checkmark
& \best{100.00}
& 99.67 & \best{100.00} & 99.66 \\

Idiosyncrasies~\citep{idiosyncrasies}
& \checkmark
& \best{100.00}
& 99.65 & \best{100.00} & 99.21 \\

\midrule
\rowcolor{gray!10}
\mname{} (ours)
& \checkmark
& \best{100.00}
& \best{99.96} & \best{100.00} & \best{99.75} \\

\bottomrule
\end{tabular}
\end{table}

\subsection{Ablation and Scaling Analysis}
\label{sec:analysis}

\begin{wraptable}{r}{0.4\textwidth}
\vspace{-9pt}
\centering
\footnotesize
\setlength{\tabcolsep}{4.2pt}
\renewcommand{\arraystretch}{1.05}
\caption{
\textbf{Component analysis.}
Accuracy (\%, $\uparrow$) under the two version-related shifts.
}
\label{tab:components}

\begin{tabular}{
@{}l
@{\hspace{6pt}}
l
@{\hspace{6pt}}
cc@{}
}
\toprule
Up-projection & Scoring & Version & Joint \\
\midrule
Shared $B$       & Mean    & 76.06 & 66.11 \\
Per-source $B_k$ & Mean    & 78.24 & 66.95 \\
\midrule
\rowcolor{gray!10}
Per-source $B_k$ & Clipped & \best{82.79} & \best{72.07} \\
\bottomrule
\end{tabular}
\vspace{-6pt}
\end{wraptable}

\paragraph{Source modeling and likelihood aggregation.}

We analyze the two main design choices of \mname{}, source-specific adaptation and clipped likelihood aggregation.
Source-specific \(B_k\) gives each source condition its own adaptation parameters, while the clipping is a technique to carefully aggregate the token-level likelihoods.
As shown in Table~\ref{tab:components}, both components improve performance when the source-model version changes.
Particularly, the clipping leads to larger gains, improving Version and Joint shift by 4.55pp and 5.12pp, respectively.
Table~\ref{tab:components_full} in Appendix~\ref{app:exp_analysis} includes the results on ID and Domain shift, showing that the performance differs by less than 0.7pp on the two settings.

\begin{wraptable}{r}{0.45\textwidth}
\vspace{-8pt}
\centering
\footnotesize
\setlength{\tabcolsep}{4.2pt}
\renewcommand{\arraystretch}{1.05}
\caption{
\textbf{Effect of backbone family.}
Accuracy (\%, $\uparrow$) under three different backbones.
}
\label{tab:backbone_family_main}

\begin{tabular}{
@{}l
@{\hspace{6pt}}
ccc@{}
}
\toprule
Backbone & Domain & Version & Joint \\
\midrule
\rowcolor{gray!10}
Qwen2.5-7B & 97.59 & 82.79 & 72.07 \\
OLMo-2-7B  & 97.05 & \best{83.75} & 72.99 \\
Qwen3-8B   & \best{98.09} & 83.40 & \best{75.22} \\
\bottomrule
\end{tabular}
\vspace{-6pt}
\end{wraptable}

\paragraph{Effect of backbone model size.}

\begin{figure}[t]
\centering
\begin{subfigure}[b]{0.5\linewidth}
\centering
\begin{tikzpicture}
\begin{axis}[
  width=4.6cm, height=3.6cm, scale only axis,
  xmin=0.75, xmax=4.25, xtick={1,2,3,4}, xticklabels={0.5B,1.5B,3B,7B},
  xlabel={Qwen2.5 backbone size}, ylabel={Accuracy (\%)},
  ymin=60, ymax=100, ytick={60,70,80,90,100},
  axis lines*=left, xmajorgrids=true, ymajorgrids=true,
  grid style={black!22,line width=0.4pt},
  axis line style={black!55}, tick style={black!55}, tick align=outside,
  tick label style={font=\footnotesize}, label style={font=\footnotesize},
  mark size=1.8pt, line width=1.1pt, clip=false,
  legend style={font=\scriptsize, at={(0.5,1.03)}, anchor=south, draw=none, fill=none,
                legend columns=2, /tikz/every even column/.append style={column sep=6pt}},
  legend cell align=left,
]
\addplot[bbGray,mark=*,line width=0.9pt] coordinates {(1,98.32)(2,99.01)(3,99.38)(4,99.64)};
\addplot[bbNavy!45,mark=square*]         coordinates {(1,88.19)(2,96.17)(3,94.59)(4,97.59)};
\addplot[bbNavy!72,mark=triangle*]       coordinates {(1,78.33)(2,81.00)(3,82.18)(4,82.79)};
\addplot[bbNavy,mark=diamond*]           coordinates {(1,64.98)(2,70.28)(3,71.21)(4,72.07)};
\legend{ID,Domain Shift,Version Shift,Joint Shift}
\end{axis}
\end{tikzpicture}
\caption{Backbone size}
\label{fig:backbone_scale_a}
\end{subfigure}%
\begin{subfigure}[b]{0.5\linewidth}
\centering
\begin{tikzpicture}
\begin{axis}[
  width=4.9cm, height=3.6cm, scale only axis,
  xmin=0.6, xmax=6.4,
  xtick={1,2,3,4,5,6}, xticklabels={1k,3k,5k,10k,15k,30k},
  xlabel={Training data size per source}, ylabel={Accuracy (\%)},
  ymin=55, ymax=85, ytick={55,60,65,70,75,80,85},
  axis lines*=left, xmajorgrids=true, ymajorgrids=true,
  grid style={black!22,line width=0.4pt},
  axis line style={black!55}, tick style={black!55}, tick align=outside,
  tick label style={font=\footnotesize}, label style={font=\footnotesize},
  mark size=1.7pt, line width=1.1pt, clip=false,
  legend style={font=\scriptsize, at={(0.5,1.03)}, anchor=south, draw=none, fill=none,
                legend columns=2, /tikz/every even column/.append style={column sep=6pt}},
  legend cell align=left,
]
\addplot[bbNavy,mark=diamond*] coordinates
  {(1,80.10)(2,81.05)(3,81.95)(4,82.79)(5,83.42)(6,84.36)};
\addplot[bbNavy,dashed,mark=o,mark options={solid}] coordinates
  {(1,59.82)(2,66.87)(3,66.72)(4,67.43)(5,69.38)(6,72.78)};
\addplot[black!45,mark=diamond*] coordinates
  {(1,67.20)(2,69.01)(3,69.32)(4,72.07)(5,73.08)(6,74.30)};
\addplot[black!45,dashed,mark=o,mark options={solid}] coordinates
  {(1,55.96)(2,62.32)(3,63.46)(4,63.81)(5,64.31)(6,66.15)};
\legend{\mname{} (Version),DA-MTL (Version),\mname{} (Joint),DA-MTL (Joint)}
\draw[bbNavy!70,densely dotted,line width=0.9pt] (axis cs:0.6,72.78) -- (axis cs:6.4,72.78);
\draw[black!45,densely dotted,line width=0.9pt] (axis cs:0.6,66.15) -- (axis cs:6.4,66.15);
\draw[black!40,line width=0.4pt,fill=white] (rel axis cs:0.37,0.02) rectangle (rel axis cs:0.99,0.13);
\draw[black!60,densely dotted,line width=0.9pt] (rel axis cs:0.40,0.075) -- (rel axis cs:0.49,0.075);
\node[anchor=west,font=\scriptsize,inner sep=1pt] at (rel axis cs:0.50,0.075) {DA-MTL at 30k data};
\end{axis}
\end{tikzpicture}
\caption{Training data size}
\label{fig:data_scale_main}
\end{subfigure}
\caption{
\textbf{Effect of backbone and training data size.}
(a) Accuracy (\%, $\uparrow$) of \mname{} with four Qwen2.5 backbone sizes, each trained on 10,000 texts per source family.
(b) Accuracy (\%, $\uparrow$) on Version Shift and Joint Shift against the number of training texts per source, compared with DA-MTL. The dotted lines mark DA-MTL on the full split of 30k texts.
}
\label{fig:backbone_scale}
\end{figure}
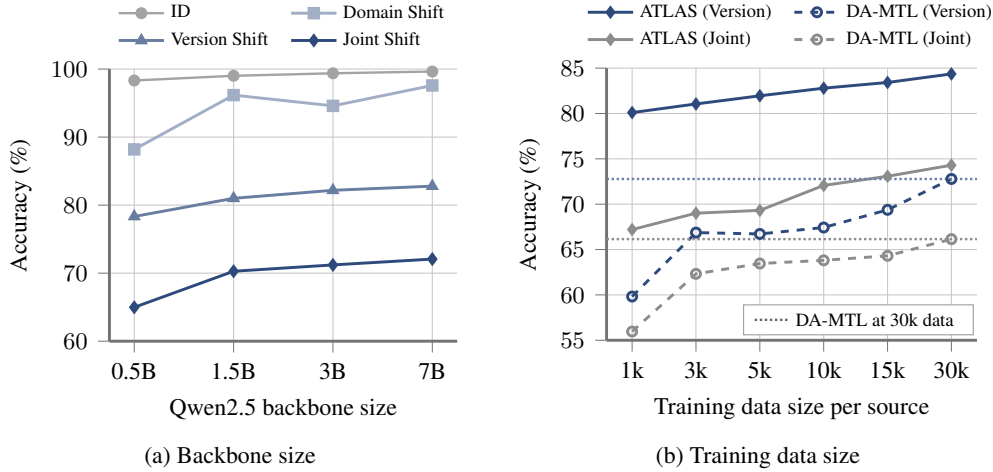

We vary the Qwen2.5 backbone from 0.5B to 7B parameters and compare the results.
As shown in Figure~\ref{fig:backbone_scale}(a), larger backbones generally improve attribution performance, with substantially larger gains under OOD evaluation than under ID.
We further replace the backbone with OLMo-2-7B and Qwen3-8B.
As shown in Table~\ref{tab:backbone_family_main}, accuracy stays within about 1pp of Qwen2.5-7B on Domain and Version shift, and Qwen3-8B improves Joint shift to 75.22\%, so the gains of \mname{} do not depend on a particular backbone.
Table~\ref{tab:backbone_family} in Appendix~\ref{app:exp_analysis} additionally reports accuracy and macro-F1 on the ID test data.

\paragraph{Data efficiency of \mname{}.}
We also vary the number of training texts per source family from 1,000 to the full split, and train the four encoder-tuning baselines on subsets of the same sizes.
As shown in Figure~\ref{fig:backbone_scale} (b), \mname{} is markedly more data-efficient.
Notably, with only 1,000 texts per source, it already outperforms DA-MTL trained on the full split on both Version shift (80.10\% vs. 72.78\%) and Joint shift (67.20\% vs. 66.15\%).
Appendix~\ref{app:exp_analysis} reports the results for all four settings, showing similar trends.

\section{Beyond Attribution: Distilled and Deployed Models}
\label{sec:additional_eval}

Beyond text-level attribution, we ask whether the source-conditioned likelihoods learned by \mname{} capture family-specific generation patterns that persist after distillation.
In specific, a student trained on the outputs of a teacher may retain such characteristics in its generations.
We first evaluate this question on distilled models with explicitly reported teachers, and then extend the analysis to recently deployed models for which no teacher, or even the use of distillation, is publicly established.

\paragraph{Setup.}
From the \benchname{} training splits, we here use only the data from GPT~\citep{gpt5}, Claude~\citep{claude}, and Gemini~\citep{gemini} for training,
since they are among the most widely used frontier model families and therefore plausible sources of teacher signals~\citep{aiindex}.
For validation, we use six publicly available distilled models whose teachers are explicitly reported and belong to one of these three families.
We consider two students for each teacher, built on different base models (see the distilled model details in Table~\ref{tab:distill_validation}).

Beyond these models with known distillation histories, we also examine seven recent deployed models from DeepSeek~\citep{deepseek}, GLM~\citep{glm}, MiniMax~\citep{minimax}, and Kimi~\citep{kimi}.
For these models, we do not assume that they were distilled from the three source families, or that distillation was used at all.
Instead, we use them as an exploratory setting to examine which, if any, known source family their generations most closely resemble.
Appendix~\ref{app:exp_distillation} gives further details about how we generate texts using the models.

\paragraph{Inference.}

We compute the mean token log-likelihood $s_k(y):=\frac{1}{n}\sum_{t=1}^{n}\ell_{k,t}$ under each source condition $k$ and subtract the corresponding likelihood $s_0(y)$ under the frozen backbone, yielding $\tilde{s}_k(y)=s_k(y)-s_0(y)$.
We then standardize $\tilde{s}_k(y)$, yielding a source-specific score $z_k(y)=(\tilde{s}_k(y)-\mu_k)/\sigma_k$, where $\mu_k$ and $\sigma_k$ are the mean and standard deviation of $\tilde{s}_k$ over reference texts from Mistral and Llama, which lie outside the three candidate families.
Here, $z_k(y)$ measures how strong the condition $k$ explains the text relative to unrelated families.
Each text is attributed to the source with the largest $z_k(y)$, and we report the attribution share for each model.
We additionally report the rejection rate, defined as the fraction of texts for which no source condition improves over the frozen backbone, i.e., $\max_k \tilde{s}_k(y)<0$.

\paragraph{Results.}

\pgfplotsset{distside/.style={
  scale only axis, clip=false,
  point meta=explicit symbolic,
  every axis plot/.append style={draw=white,line width=0.5pt},
  tick label style={font=\footnotesize}, label style={font=\footnotesize},
  grid style={black!12,line width=0.3pt},
  axis line style={black!55}, tick style={black!55},
}}
\newcommand{\distcolgrp}[4]{%
  \draw[black!35,line width=0.5pt] ([yshift=-0.34cm]axis cs:#1-0.45,0) -- ([yshift=-0.34cm]axis cs:#2+0.45,0);
  \node[anchor=north,font=\footnotesize\bfseries,text=#4]
    at ([yshift=-0.38cm]axis cs:{(#1+#2)/2},0) {#3};}
\newcommand{\distcell}[4]{%
  \fill[#3!#4] (3.0+#1*1.1,#2-0.44) rectangle (3.0+#1*1.1+1.06,#2);
  \node[font=\scriptsize,text=black!80] at (3.0+#1*1.1+0.53,#2-0.22) {#4};}
\newcommand{\distnone}[2]{%
  \fill[bbGray!#2] (6.45,#1-0.44) rectangle (7.51,#1);
  \node[font=\scriptsize,text=black!80] at (6.98,#1-0.22) {#2};}
\newcommand{\distrow}[5]{%
  \node[anchor=east,font=\footnotesize] at (2.92,#1-0.22) {#2};
  \distcell{0}{#1}{famG}{#3}\distcell{1}{#1}{famC}{#4}\distcell{2}{#1}{famM}{#5}}
\newcommand{\distteach}[4]{%
  \draw[black!35,line width=0.5pt] (1.02,#1-0.03) -- (1.02,#1-#2*0.50+0.02);
  \node[anchor=east,font=\footnotesize\bfseries,text=#4] at (0.94,#1-#2*0.25) {#3};}

\begin{figure}[t]
\centering
\begin{tikzpicture}
\distrow{4.2}{GPT-distilled 1}{93.6}{5.3}{1.1}
\distnone{4.2}{10.6}
\distrow{3.74}{GPT-distilled 2}{87.3}{11.9}{0.8}
\distnone{3.74}{30.2}
\distrow{3.12}{Claude-distilled 1}{5.8}{92.9}{1.3}
\distnone{3.12}{1.8}
\distrow{2.66}{Claude-distilled 2}{23.0}{76.4}{0.6}
\distnone{2.66}{8.0}
\distrow{2.04}{Gemini-distilled 1}{13.4}{10.6}{76.0}
\distnone{2.04}{0.6}
\distrow{1.58}{Gemini-distilled 2}{30.6}{6.1}{63.3}
\distnone{1.58}{1.5}
\draw[black!45,densely dashed,line width=0.5pt] (0.4,1.05) -- (7.51,1.05);
\distrow{0.96}{\textit{Comparison 1}}{40.6}{23.2}{36.2}
\distnone{0.96}{68.2}
\distrow{0.50}{\textit{Comparison 2}}{37.7}{18.7}{43.7}
\distnone{0.50}{74.6}
\foreach \x in {3.53,4.63,5.73,6.98}{\draw[black!35,line width=0.5pt] (\x-0.48,-0.34) -- (\x+0.48,-0.34);}
\node[font=\footnotesize\bfseries,text=famG!75!black,anchor=north] at (3.53,-0.38) {GPT};
\node[font=\footnotesize\bfseries,text=famC!75!black,anchor=north] at (4.63,-0.38) {Claude};
\node[font=\footnotesize\bfseries,text=famM!70!black,anchor=north] at (5.73,-0.38) {Gemini};
\node[font=\footnotesize\bfseries,text=black!60,anchor=north] at (6.98,-0.38) {None};
\node[anchor=south west,font=\footnotesize\bfseries] at (-0.2,4.34) {(a) Distilled models (with known teachers)};
\begin{axis}[distside,
  at={(8.45cm,0)}, anchor=south west,
  ybar stacked, bar width=13pt, width=4.85cm, height=4.2cm,
  ymin=0, ymax=100, ytick={0,25,50,75,100}, ymajorgrids=true,
  xmin=-0.6, xmax=8.4,
  axis x line*=bottom, axis y line*=left, x axis line style={draw=none}, xtick style={draw=none},
  xtick={0,1,2.6,3.6,5.2,6.2,7.8},
  xticklabels={V4-Pro,V3.2,5.2,5.1,M3,M2.7,K2.6},
  xticklabel style={font=\tiny,anchor=north,inner sep=1pt,yshift=-1pt},
  ylabel={Share (\%)}, ylabel style={yshift=-8pt},
  nodes near coords={\ifdim\pgfplotspointmeta pt>12pt\pgfplotspointmeta\fi},
  nodes near coords style={font=\tiny,text=black!80,anchor=center},
  legend to name=distlegF, legend columns=3, legend cell align=left,
  legend style={font=\tiny, draw=black!40, fill=white, line width=0.4pt, inner sep=2pt,
                /tikz/every even column/.append style={column sep=4pt}},
  legend image code/.code={\fill[#1] (0cm,-0.07cm) rectangle (0.22cm,0.07cm);},
]
\addplot[fill=famG] coordinates {(0,90.6)[90.6] (1,41.5)[41.5] (2.6,15.2)[15.2] (3.6,31.5)[31.5] (5.2,56.4)[56.4] (6.2,6.2)[6.2] (7.8,14.8)[14.8]};
\addplot[fill=famC] coordinates {(0,9.3)[9.3] (1,27.8)[27.8] (2.6,84.7)[84.7] (3.6,63.6)[63.6] (5.2,43.6)[43.6] (6.2,93.8)[93.8] (7.8,85.2)[85.2]};
\addplot[fill=famM] coordinates {(0,0.1)[0.1] (1,30.6)[30.6] (2.6,0.1)[0.1] (3.6,4.9)[4.9] (5.2,0.1)[0.1] (6.2,0.0)[0.0] (7.8,0.0)[0.0]};
\legend{GPT,Claude,Gemini}
\distcolgrp{0}{1}{DeepSeek}{black!80}
\distcolgrp{2.6}{3.6}{GLM}{black!80}
\distcolgrp{5.2}{6.2}{MiniMax}{black!80}
\distcolgrp{7.8}{7.8}{Kimi}{black!80}
\node[anchor=south west,font=\footnotesize\bfseries] at ([yshift=4pt]rel axis cs:0,1) {(b) Deployed models (resemblance)};
\end{axis}
\node[anchor=north west,inner sep=0pt] at (-0.25,-0.34) {\pgfplotslegendfromname{distlegF}};
\end{tikzpicture}
\caption{
\textbf{Attribution of distilled and deployed models.}
Share of each model's texts attributed to each candidate family, in percent.
(a) Distilled models with known teachers: \emph{None} is the share of texts that no condition explains better than the frozen backbone.
(b) Seven recent releases of widely deployed models: segments under $12\%$ are left unlabeled.
}
\label{fig:distill}
\end{figure}
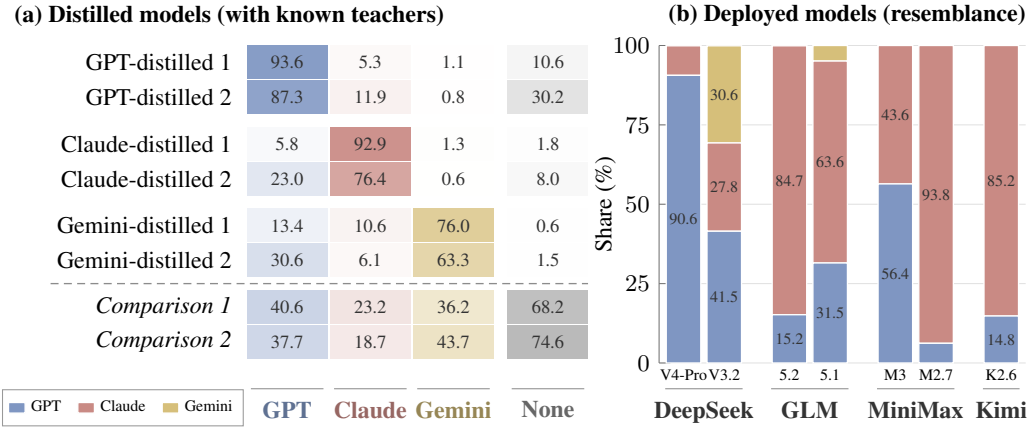

As the left panel (a) of \cref{fig:distill} shows, using \mname{}, we successfully attribute all six distilled models to their true known teachers, with scores from 63.3\% to 93.6\%.
It is noteworthy that the two GPT-distilled students are both attributed to GPT, even though they are built on different base models, Qwen3-8B and Llama-3.1-8B, providing evidence that the learned source conditions capture common generation patterns of the teacher model family rather than only the student's base model.
Table~\ref{tab:distill} in Appendix~\ref{app:exp_distillation} presents the detailed results.

The right panel (b) of \cref{fig:distill} shows that, among the deployed models with no reported teacher, four models are mostly attributed to a single source family, at levels comparable to the distilled models with known teachers.
That is, MiniMax-M2.7 Turbo, Kimi-K2.6, and GLM 5.2 most resemble Claude in 93.8\%, 85.2\%, and 84.7\% of their texts, respectively, while DeepSeek-V4-Pro most resembles GPT in 90.6\%.
Appendix~\ref{app:exp_distillation} extends this analysis to five additional deployed models, Grok-4.7~\citep{grok} and four open-weight models, A.X-K2~\citep{axk2}, K-EXAONE-2.0~\citep{exaone}, Motif-3~\citep{motif3}, and Solar-Open2~\citep{solar}, all of which are most attributed to GPT.

The rejection rate ranges from $0.6\%$ to $30.2\%$ for the validated distilled models and is at most $0.4\%$ for the seven deployed models, but rises to $68.2\%$ and $74.6\%$ for two comparison models, Falcon3-7B-Instruct and Granite-3.0-8B-Instruct, with no reported distillation from three families.
Overall, these results show that \mname{} not only recovers the (known) teachers of distilled models, but also characterizes which model family a given deployed model most closely resembles.

\section{Conclusion and Discussion}
\label{sec:conclusion}

In this work, we developed a new method \mname{} for model authorship attribution (MAA) through source-conditioned text modeling rather than direct source prediction.
\mname{} learns how each source family generates text and performs attribution from the resulting likelihood evidence.
On \benchname{} and OpenTuringBench, \mname{} consistently improves attribution under distribution shift compared to existing baselines, especially for the most challenging OOD setting when both the generation domain and the source-model version change at test time.
The same source-conditioned likelihoods can also be applied to distinguish human-written text from MGT.
They further identify the known teachers of distilled models and indicate which family deployed models most resemble.

This work focuses on attribution among a fixed set of candidate source families under the domains and generation conditions covered by \benchname{}.
Extending attribution to previously unseen source families, particularly without full retraining, remains an important direction for future work.
Another promising extension is mixed-authorship attribution for text jointly written by humans and models, where the task shifts from assigning a single source to identifying which source contributed each part of the text.

\bibliographystyle{preprint}
\bibliography{references}

\begin{thebibliography}{57}
\providecommand{\natexlab}[1]{#1}
\providecommand{\url}[1]{\texttt{#1}}
\expandafter\ifx\csname urlstyle\endcsname\relax
  \providecommand{\doi}[1]{doi: #1}\else
  \providecommand{\doi}{doi: \begingroup \urlstyle{rm}\Url}\fi

\bibitem[Anthropic(2026{\natexlab{a}})]{claude}
Anthropic.
\newblock System card: Claude opus 5, 2026{\natexlab{a}}.
\newblock URL \url{https://www-cdn.anthropic.com/ceaf5c7ff2783855203fde8208ec311252dced5b/Claude%20Opus%205%20System%20Card.pdf}.

\bibitem[Anthropic(2026{\natexlab{b}})]{claude_misuse}
Anthropic.
\newblock Detecting and countering misuse of {AI}: September 2026, September 2026{\natexlab{b}}.
\newblock URL \url{https://www-cdn.anthropic.com/e50be2e51e7695dc4b1366a37a245a597377d3b5/Anthropic-Detecting-and-countering-091026.pdf}.

\bibitem[Baek et~al.(2026)Baek, Arya, Kim, Song, Han, Yang, Eun, Kim, Park, Wee, Hong, Park, Kim, Kang, Kim, Kim, Lee, Lee, Jo, Hong, Choi, Cho, Ok, Sek, Cho, Kim, Kim, Park, Park, Yi, Yoon, Lee, Cheon, Kim, Lee, Kim, Jang, Ra, jin Han, Kim, Kim, Kang, and Lee]{axk2}
Cheolseung Baek, Dhammiko Arya, Eunki Kim, Gun Song, Gyoungeun Han, Hyunho Yang, Hyunjun Eun, Jin Kim, Junyoung Park, Juyun Wee, Minki Hong, Minkyung Park, Minsang Kim, Minsoo Kang, SaeRom Kim, Sangjin Kim, Sangyeol Lee, Seojin Lee, Seokhwan Jo, Seokyoung Hong, Seongho Choi, Seonghye Cho, Seongmin Ok, Sereimony Sek, Seungmo Cho, Seungsik Kim, Singon Kim, Sohee Park, Sooyeon Park, Subin Yi, Sungbin Yoon, Sungeun Lee, Sung~Jun Cheon, Sungwan Kim, Sunwoo Lee, Tae~Yoon Kim, Wonbeom Jang, Yohan Ra, Yong jin Han, Youngjin Kim, Youngrang Kim, Yujin Kang, and Yujin Lee.
\newblock {A.X K2} technical report, 2026.
\newblock URL \url{https://arxiv.org/abs/2608.30181}.

\bibitem[Bao et~al.(2024)Bao, Zhao, Teng, Yang, and Zhang]{fastdetectgpt}
Guangsheng Bao, Yanbin Zhao, Zhiyang Teng, Linyi Yang, and Yue Zhang.
\newblock Fast-detect{GPT}: Efficient zero-shot detection of machine-generated text via conditional probability curvature.
\newblock In \emph{The Twelfth International Conference on Learning Representations}, 2024.
\newblock URL \url{https://openreview.net/forum?id=Bpcgcr8E8Z}.

\bibitem[BehnamGhader et~al.(2024)BehnamGhader, Adlakha, Mosbach, Bahdanau, Chapados, and Reddy]{llm2vec}
Parishad BehnamGhader, Vaibhav Adlakha, Marius Mosbach, Dzmitry Bahdanau, Nicolas Chapados, and Siva Reddy.
\newblock {LLM}2vec: Large language models are secretly powerful text encoders.
\newblock In \emph{First Conference on Language Modeling}, 2024.
\newblock URL \url{https://openreview.net/forum?id=IW1PR7vEBf}.

\bibitem[Chen et~al.(2026)Chen, Li, Zhou, Gong, Jiang, Dan, Zhang, Yu, Wang, Ma, Zhong, Zhu, Xiao, Yang, Du, Zhang, Zhang, Huang, Zhang, Du, Zhao, Guo, Chen, Ding, Sun, Li, Zhang, Yang, Yu, Zheng, Zheng, Li, Zhu, Zhou, Zhang, Ding, Zhang, Sun, Lyu, Lu, Wang, Shi, Li, Chen, Zhang, Zhuang, Cai, Pan, Li, Song, Zhang, Wang, Gu, Zhu, Dong, Li, Zhang, Zhuang, Tian, Liu, Hu, Tao, Zhang, Ruan, Xu, Yan, Liu, He, Xu, Ji, Yang, Xiao, Duan, Li, Han, Ruan, Yuan, Yu, Feng, Mo, Li, Du, Bao, Yang, Zhou, Loki, Chen, Zeng, Li, Zhong, Tao, Chi, Lin, Hu, Chen, Zhu, Gao, Gao, Li, Li, Zhao, Ren, Ren, Xu, Ren, Li, Wang, Chen, Zeng, Tian, Guo, Dong, Leng, Zhang, Liu, Chen, Chen, Jia, Yao, Zhao, Yu, Li, Pan, Zhu, Li, Xie, Qin, Li, Liang, Liu, Xu, Li, Chen, Cheng, Zhang, Chen, Zhao, Chen, Song, Wang, Luo, Luo, Su, Li, Han, Wu, Song, Han, Guan, Lu, Zou, Lai, Li, Shen, Gong, Ma, Jiao, Wang, Xu, Wang, Tang, Chen, Wang, Qiu, Shi, Guo, Huang, Wang, Hu, Gao, Zhang, Li, Ying, Zhang, Wang, Song, Yang, Meng, Miao, Li, Liu, Hu, Huang, Li, Huang, Zhang, Hong, Xie, Zhang, Liao, Shi, Wenren, Li, Li, Luo, Jin, Sun, Zhou, Su, Li, Zhu, Peng, Fan, Zhang, Xu, Lv, Xu, He, He, Li, Gao, Wu, Song, Zhou, Sun, Huang, Chen, and Ge]{minimax}
Aili Chen, Aonian Li, Baichuan Zhou, Bangwei Gong, Binyang Jiang, Boji Dan, Changhao Zhang, Changqing Yu, Chao Wang, Cheng Ma, Cheng Zhong, Cheng Zhu, Chengjun Xiao, Chengyi Yang, Chengyu Du, Chenyang Zhang, Chi Zhang, Chuangyi Huang, Chunhao Zhang, Chunhui Du, Chunyu Zhao, Congchao Guo, Da~Chen, Deming Ding, Dianjun Sun, Dong Li, Dongyu Zhang, Enhui Yang, Fei Yu, Guang Zheng, Guodong Zheng, Guohong Li, Haichao Zhu, Haigang Zhou, Haimo Zhang, Han Ding, Hao Zhang, Haohai Sun, Haolin Lyu, Haonan Lu, Haoyu Wang, Huajie Shi, Huiyang Li, Jiacheng Chen, Jian Zhang, Jiaqi Zhuang, Jiaren Cai, Jiaxin Pan, Jiayao Li, Jiayuan Song, Jichuan Zhang, Jie Wang, Jihao Gu, Jin Zhu, Jingwei Dong, Jingyang Li, Jingyu Zhang, Jingze Zhuang, Jinhao Tian, Jinli Liu, Jinyi Hu, Jun Tao, Jun Zhang, Junbin Ruan, Junhao Xu, Junjie Yan, Junteng Liu, Junxian He, Kang Xu, Ke~Ji, Ke~Yang, Kecheng Xiao, Keyu Duan, Keyu Li, Le~Han, Letian Ruan, Li~Yuan, Lianfei Yu, Liheng Feng, Lijie Mo, Lin Li, Linge Du, Lingye Bao, Lingyu Yang, Lingyuan Zhou, Loki, Lu~Chen, Lunbin Zeng, Ming Li, Ming Zhong, Mingliang Tao, Mingyuan Chi, Mujie Lin, Nan Hu, Ningxin Chen, Peiyin Zhu, Peng Gao, Pengcheng Gao, Pengfei Li, Penglin Li, Pengyu Zhao, Qibin Ren, Qibing Ren, Qidi Xu, Qihan Ren, Qile Li, Qin Wang, Quanliang Chen, Qunhong Zeng, Rong Tian, Rongxin Guo, Rui Dong, Ruitao Leng, Ruize Zhang, Shanqi Liu, Shaoxiang Chen, Shaoyu Chen, Sheng Jia, Shun Yao, Shuoran Zhao, Shuqi Yu, Sichen Li, Sicheng Pan, Songquan Zhu, Tengfei Li, Tian Xie, Tiancheng Qin, Tianle Li, Tianrun Liang, Wei Liu, Weiqi Xu, Weitao Li, Weixiang Chen, Weiyu Cheng, Weiyu Zhang, Wenhu Chen, Wenqian Zhao, Xiancai Chen, Xiangjun Song, Xiangyuan Wang, Xianzhen Luo, Xiao Luo, Xiao Su, Xiaobo Li, Xiaodong Han, Xiaojie Wu, Xihao Song, Xingyi Han, Xinyu Guan, Xuan Lu, Xun Zou, Xunhao Lai, Xutong Li, Xuyang Shen, Yan Gong, Yan Ma, Yang Jiao, Yang Wang, Yang Xu, Yangsen Wang, Ye~Tang, Yicheng Chen, Yihang Wang, Yinran Qiu, Yiqi Shi, Yiting Guo, Yiwen Huang, Yixuan Wang, Yongyi Hu, Yu~Gao, Yu~Zhang, Yuan Li, Yuanxiang Ying, Yuanzhen Zhang, Yubo Wang, Yuchen Song, Yufeng Yang, Yuhang Meng, Yuhang Miao, Yuhao Li, Yujie Liu, Yulin Hu, Yunan Huang, Yunji Li, Yunyi Huang, Yusen Zhang, Yusu Hong, Yutao Xie, Yutong Zhang, Yuwen Liao, Yuxuan Shi, Yuze Wenren, Zebin Li, Zehan Li, Zejian Luo, Zeyu Jin, Zeyuan Sun, Zhanpeng Zhou, Zhaochen Su, Zhendong Li, Zhengmao Zhu, Zhengyuan Peng, Zhenhua Fan, Zhi Zhang, Zhichao Xu, Zhiheng Lv, Zhikang Xu, Zhitao He, Zhiwei He, Zhongyuan Li, Zibo Gao, Zijia Wu, Zijian Song, Zijian Zhou, Zijun Sun, Zishan Huang, Ziying Chen, and Ziyue Ge.
\newblock The minimax-m2 series: Mini activations unleashing max real-world intelligence, 2026.
\newblock URL \url{https://arxiv.org/abs/2605.26494}.

\bibitem[Chen et~al.(2023{\natexlab{a}})Chen, Kang, Zhai, Li, Singh, and Raj]{gptsentinel}
Yutian Chen, Hao Kang, Vivian Zhai, Liangze Li, Rita Singh, and Bhiksha Raj.
\newblock Gpt-sentinel: Distinguishing human and chatgpt generated content, 2023{\natexlab{a}}.
\newblock URL \url{https://arxiv.org/abs/2305.07969}.

\bibitem[Chen et~al.(2023{\natexlab{b}})Chen, Kang, Zhai, Li, Singh, and Raj]{t5sentinel}
Yutian Chen, Hao Kang, Yiyan Zhai, Liangze Li, Rita Singh, and Bhiksha Raj.
\newblock Token prediction as implicit classification to identify {LLM}-generated text.
\newblock In Houda Bouamor, Juan Pino, and Kalika Bali (eds.), \emph{Proceedings of the 2023 Conference on Empirical Methods in Natural Language Processing}, pp.\  13112--13120, Singapore, December 2023{\natexlab{b}}. Association for Computational Linguistics.
\newblock \doi{10.18653/v1/2023.emnlp-main.810}.
\newblock URL \url{https://aclanthology.org/2023.emnlp-main.810/}.

\bibitem[Choi et~al.(2026)Choi, Choi, Chun, Hong, Hwang, Jeon, Jo, Jo, Jo, Jung, Kim, Kim, Kim, Kim, Kim, Kim, Kim, Kim, Ko, Lee, Lee, Lee, Lee, Lee, Lee, Lee, Park, Park, Ryoo, Seo, Seo, Song, Yang, Yeen, Choi, Choi, Chun, Ham, Hong, Im, Jeon, Jo, Jo, Jo, Jung, Kang, Kim, Kim, Kim, Kim, Kim, Kim, Kim, Lee, Lee, Lee, Lee, Lee, Lee, Lee, Lim, Lim, Moon, Mun, Park, Park, Park, Seo, Shin, Son, Son, Um, Yang, Yea, Yi, Yoo, and Yoon]{exaone}
Eunbi Choi, Kibong Choi, Sehyun Chun, Seokhee Hong, Junwon Hwang, Hyojin Jeon, Ahra Jo, Hyunjik Jo, Yeonsik Jo, Minhyeok Jung, Doyoung Kim, Heegyu Kim, Joonkee Kim, Seonghwan Kim, Soyeon Kim, Sunkyoung Kim, Yireun Kim, Yongil Kim, Byungoh Ko, Changhun Lee, Dohaeng Lee, Haeju Lee, Jinsik Lee, Kyungmin Lee, Minwoo Lee, Wonkee Lee, Sangha Park, Sungjune Park, Kwangrok Ryoo, Kijung Seo, Minju Seo, Yongwoo Song, Sejong Yang, Heuiyeen Yeen, Stanley~Jungkyu Choi, Yemuk Choi, Yongchan Chun, Jiwon Ham, Dasol Hong, Sujeong Im, Kijeong Jeon, Gerrard~Jeongwon Jo, Hyeongjun Jo, Yujin Jo, Jiyeon Jung, Naeun Kang, Daeseong Kim, Euisoon Kim, Hayeon Kim, Hyosang Kim, Myoungshin Kim, Unsol Kim, Youchul Kim, Chaeeun Lee, ChaeYoon Lee, Edward~Hwayoung Lee, Honglak Lee, Hwansoo Lee, Minkyung Lee, Sangeun Lee, Solji Lim, Woohyung Lim, Chanwoo Moon, Jueun Mun, Jimin Park, Seojeong Park, Yongmin Park, Hyerin Seo, Donghyeon Shin, Donghyun Son, Eunyong Son, Kaehyun Um, Sihoon Yang, Chang~En Yea, Sihyuk Yi, Kyungjae Yoo, and Chansik Yoon.
\newblock {K-EXAONE 2.0} technical report, 2026.
\newblock URL \url{https://arxiv.org/abs/2608.04505}.

\bibitem[Clement et~al.(2019)Clement, Bierbaum, O'Keeffe, and Alemi]{arxiv}
Colin~B. Clement, Matthew Bierbaum, Kevin~P. O'Keeffe, and Alexander~A. Alemi.
\newblock On the use of arxiv as a dataset, 2019.
\newblock URL \url{https://arxiv.org/abs/1905.00075}.

\bibitem[DeepSeek-AI et~al.(2026)DeepSeek-AI, Xu, Lin, Xue, Wang, Xu, Wu, Zhang, Lin, Dong, Ling, Lu, Zhao, Deng, Hou, Xu, Shao, Ruan, Sun, Dai, Guo, Yang, Chen, Li, Ji, Li, Wei, Lin, Yuan, Xia, Dai, Hao, Chen, Cao, Meng, Li, Yu, Zhang, Xu, Li, Liang, Zhang, Luo, Wei, Yuan, Zhang, Luo, Chen, Ji, Zhang, Ding, Tang, Cao, Gao, Qu, Zeng, Yang, Zhu, Luo, Song, Yu, Huang, Cai, Liang, Zhou, Ye, Li, Xu, Hu, Yang, Chen, Yan, Chen, Zhou, Xiang, Yuan, Cheng, Zhou, Zhu, Yu, Sun, Ran, Jiang, Qiu, Li, Zheng, Song, Dong, Gao, Guan, Zhou, Huang, Yu, Wang, Zhang, Wang, Xia, Zhang, Zhao, Guo, Luo, Ma, Zhu, Wang, Cai, Zhang, Chen, Di, Xu, Mei, Wang, Zhang, Zhang, Tang, Li, Zhou, Han, Wang, Huang, Wang, Cong, Wang, Zhang, Wang, Zhu, Li, Chen, Du, Jiang, Tian, Xu, Lu, Xu, Ge, Zhang, Pan, Wang, Chen, Yin, Xu, Shen, Zhang, Chen, Liu, Lu, Sun, Zhou, Chen, Cai, Nie, Wu, Chen, Hu, Liu, Hu, Ma, Wang, Yu, Zhou, Pan, Yu, Zhou, Ni, Yun, Jin, Pei, Ye, Lin, Ji, Cui, Yue, Yu, Wang, Zhang, Xiao, Zeng, An, Zhao, Liu, Liang, Pang, Luo, Yao, Gao, Yang, Huang, Hou, Zhang, Ma, Gao, He, Wang, Wang, Bi, Liu, Wang, Chen, Zhang, Nie, Sun, Wang, Cheng, Liu, Xie, Liu, Liu, Yu, Li, Yang, Zhang, Chen, Wang, Su, Chen, Lin, Fu, Yan, Wang, Ma, Luo, Zhang, Xu, Ma, Huang, Li, Li, Xu, Zhao, Sun, Wang, Qian, Shao, Yu, Zhang, Ding, Shi, Wu, Xiong, Ma, He, Tang, Zhou, Luo, Zhong, Piao, Wang, Zhang, Chen, Tan, Wei, Ma, Liu, Yang, Guo, Wu, Wu, Li, Cheng, Ou, Xu, Li, Wang, Yang, Xu, Wu, Meng, Zou, Zha, Xiong, Chen, Lin, Cao, Wang, Zhang, Yan, Lin, Gu, Luo, You, Liu, Zhou, Zhou, Huang, Wu, Wang, Zhao, Ren, Zhang, Sha, Fu, Ju, Xu, Xie, Zhang, Gao, Hao, Gou, Ma, Yan, Shao, Huang, Chen, Wu, Ren, Wu, Li, Zhang, Xu, Wang, Qu, Gu, Zhu, Li, Zhang, Xie, Gao, Wan, Pan, and Yao]{deepseek}
DeepSeek-AI, Anyi Xu, Bangcai Lin, Bing Xue, Bingxuan Wang, Bingzheng Xu, Bochao Wu, Bowei Zhang, Chaofan Lin, Chen Dong, Chenchen Ling, Chengda Lu, Chenggang Zhao, Chengqi Deng, Chengyu Hou, Chenhao Xu, Chenze Shao, Chong Ruan, Conner Sun, Damai Dai, Daya Guo, Dejian Yang, Deli Chen, Donghao Li, Dongjie Ji, Erhang Li, Fang Wei, Fangyun Lin, Fangzhou Yuan, Feiyu Xia, Fucong Dai, Guangbo Hao, Guanting Chen, Guoai Cao, Guolai Meng, Guowei Li, Han Yu, Han Zhang, Hanwei Xu, Hao Li, Haofen Liang, Haoling Zhang, Haoming Luo, Haoran Wei, Haotian Yuan, Haowei Zhang, Haowen Luo, Haoyu Chen, Haozhe Ji, Hengqing Zhang, Honghui Ding, Hongxuan Tang, Huanqi Cao, Huazuo Gao, Hui Qu, Hui Zeng, J~Yang, JQ~Zhu, Jia Luo, Jia Song, Jia Yu, Jialiang Huang, Jialu Cai, Jian Liang, Jiangting Zhou, Jiasheng Ye, Jiashi Li, Jiaxin Xu, Jiewen Hu, Jieyu Yang, Jin Chen, Jin Yan, Jingchang Chen, Jingli Zhou, Jingting Xiang, Jingyang Yuan, Jingyuan Cheng, Jingzi Zhou, Jinhua Zhu, Jiping Yu, Joseph Sun, Jun Ran, Junguang Jiang, Junjie Qiu, Junlong Li, Junmin Zheng, Junxiao Song, Kai Dong, Kaige Gao, Kang Guan, Kexing Zhou, Kezhao Huang, Kuai Yu, Lean Wang, Lecong Zhang, Lei Wang, Leyi Xia, Li~Zhang, Liang Zhao, Lihua Guo, Lingxiao Luo, Linwang Ma, Linyan Zhu, Litong Wang, Liyu Cai, Liyue Zhang, Longhao Chen, MS~Di, MY~Xu, Max Mei, Miaojun Wang, Mingchuan Zhang, Minghua Zhang, Minghui Tang, Mingming Li, Mingxu Zhou, Minmin Han, Ning Wang, Panpan Huang, Panpan Wang, Peixin Cong, Peiyi Wang, Peng Zhang, Qiancheng Wang, Qihao Zhu, Qingyang Li, Qinyu Chen, Qiushi Du, Qiwei Jiang, Rui Tian, Ruifan Xu, Ruijie Lu, Ruiling Xu, Ruiqi Ge, Ruisong Zhang, Ruizhe Pan, Runji Wang, Runqian Chen, Runqiu Yin, Runxin Xu, Ruomeng Shen, Ruoyu Zhang, Ruyi Chen, SH~Liu, Shanghao Lu, Shangmian Sun, Shangyan Zhou, Shanhuang Chen, Shaofei Cai, Shaoheng Nie, Shaoqing Wu, Shaoyuan Chen, Shengding Hu, Shengyu Liu, Shiqiang Hu, Shirong Ma, Shiyu Wang, Shuiping Yu, Shunfeng Zhou, Shuting Pan, Shuying Yu, Songyang Zhou, Tao Ni, Tao Yun, Tian Jin, Tian Pei, Tian Ye, Tianle Lin, Tianran Ji, Tianyi Cui, Tianyuan Yue, Tingting Yu, Tun Wang, W~Zhang, WL~Xiao, Wangding Zeng, Wei An, Weilin Zhao, Wen Liu, Wenfeng Liang, Wenjie Pang, Wenjing Luo, Wenjing Yao, Wenjun Gao, Wenkai Yang, Wenlve Huang, Wenqing Hou, Wentao Zhang, Wenting Ma, Xi~Gao, Xiang He, Xiangwen Wang, Xianzu Wang, Xiao Bi, Xiaodong Liu, Xiaohan Wang, Xiaokang Chen, Xiaokang Zhang, Xiaotao Nie, Xiaowen Sun, Xiaoxiang Wang, Xin Cheng, Xin Liu, Xin Xie, Xingchao Liu, Xingchen Liu, Xingkai Yu, Xingyou Li, Xinyu Yang, Xinyu Zhang, Xu~Chen, Xuanyu Wang, Xuecheng Su, Xueyin Chen, Xuheng Lin, Xuwei Fu, YC~Yan, YQ~Wang, YW~Ma, Yanfeng Luo, Yang Zhang, Yanhong Xu, Yanru Ma, Yanwen Huang, Yao Li, Yao Li, Yao Xu, Yao Zhao, Yaofeng Sun, Yaohui Wang, Yi~Qian, Yi~Shao, Yi~Yu, Yichao Zhang, Yifan Ding, Yifan Shi, Yijia Wu, Yiliang Xiong, Yiling Ma, Ying He, Ying Tang, Ying Zhou, Yingjia Luo, Yinmin Zhong, Yishi Piao, Yisong Wang, Yixiang Zhang, Yixiao Chen, Yixuan Tan, Yixuan Wei, Yiyang Ma, Yiyuan Liu, Yonglun Yang, Yongqiang Guo, Yongtong Wu, Yu~Wu, YuKun Li, Yuan Cheng, Yuan Ou, Yuanfan Xu, Yuanhao Li, Yuduan Wang, Yuehan Yang, Yuer Xu, Yuhan Wu, Yuhao Meng, Yuheng Zou, Yukun Zha, Yunfan Xiong, Yupeng Chen, Yuping Lin, Yuqian Cao, Yuqian Wang, Yushun Zhang, Yuting Yan, Yutong Lin, Yuxian Gu, Yuxiang Luo, Yuxiang You, Yuxuan Liu, Yuxuan Zhou, Yuyang Zhou, Yuzhen Huang, ZF~Wu, Zehao Wang, Zehua Zhao, Zehui Ren, Zekai Zhang, Zhangli Sha, Zhe Fu, Zhe Ju, Zhean Xu, Zhenda Xie, Zhengyan Zhang, Zheren Gao, Zhewen Hao, Zhibin Gou, Zhicheng Ma, Zhigang Yan, Zhihong Shao, Zhixian Huang, Zhixuan Chen, Zhiyu Wu, Zhizhou Ren, Zhongyu Wu, Zhuoshu Li, Zhuping Zhang, Zian Xu, Zihao Wang, Zihua Qu, Zihui Gu, Zijia Zhu, Zilin Li, Zipeng Zhang, Ziwei Xie, Ziyi Gao, Ziyi Wan, Zizheng Pan, and Zongqing Yao.
\newblock Deepseek-v4: Towards highly efficient million-token context intelligence, 2026.
\newblock URL \url{https://arxiv.org/abs/2606.19348}.

\bibitem[Dugan et~al.(2024)Dugan, Hwang, Trhl{\'i}k, Zhu, Ludan, Xu, Ippolito, and Callison-Burch]{raid}
Liam Dugan, Alyssa Hwang, Filip Trhl{\'i}k, Andrew Zhu, Josh~Magnus Ludan, Hainiu Xu, Daphne Ippolito, and Chris Callison-Burch.
\newblock {RAID}: A shared benchmark for robust evaluation of machine-generated text detectors.
\newblock In Lun-Wei Ku, Andre Martins, and Vivek Srikumar (eds.), \emph{Proceedings of the 62nd Annual Meeting of the Association for Computational Linguistics (Volume 1: Long Papers)}, pp.\  12463--12492, Bangkok, Thailand, August 2024. Association for Computational Linguistics.
\newblock \doi{10.18653/v1/2024.acl-long.674}.
\newblock URL \url{https://aclanthology.org/2024.acl-long.674/}.

\bibitem[Fu et~al.(2025)Fu, Guo, and Li]{detectanyllm}
Jiachen Fu, Chun-Le Guo, and Chongyi Li.
\newblock Detectanyllm: Towards generalizable and robust detection of machine-generated text across domains and models, 2025.
\newblock URL \url{https://arxiv.org/abs/2509.14268}.

\bibitem[GLM-5-Team et~al.(2026)GLM-5-Team, :, Zeng, Lv, Hou, Du, Zheng, Chen, Yin, Ge, Huang, Xie, Zhu, Yin, Wang, Pan, Zeng, Zhang, Wang, Chen, Zhang, Jiao, Guo, Wang, Du, Wu, Wang, Li, Fan, Zhong, Liu, Zhao, Du, Dong, Lu, Shuang-Li, Cao, Liu, Jiang, Chen, Zhang, Huang, Dong, Xu, Wei, An, Niu, Zhu, Wen, Cen, Bai, Qiao, Wang, Wang, Zhu, Liu, Li, Wang, Wen, Huang, Cai, Yu, Li, Hu, Zhang, Zhang, Lin, Yang, Wang, Ai, Zhu, Yi, Chen, Wen, Sun, Zhao, Hu, Zhang, Liu, Zhang, Peng, Tai, Zhang, Liu, Wang, Yan, Ge, Liu, Chu, Zhao, Wang, Zhao, Ren, Wang, Zhang, Gui, Zhao, Li, An, Li, Yuan, Du, Liu, Zhi, Duan, Zhou, Wei, Wang, Luo, Zhang, Sha, Xu, Wu, Ding, Chen, Li, Lin, Ta, Zou, Song, Yang, Tu, Yang, Wu, Zhang, Li, Li, Fan, Qin, Tian, Zhang, Yu, Liang, Kuang, Cheng, Li, Yan, Hu, Ling, Fan, Xia, Zhang, Zhang, Pan, Zou, Zhang, Liu, Wu, Li, Wang, Zhu, Tan, Zhou, Pan, Zhang, Su, Geng, Yan, Tan, Bi, Shen, Yang, Li, Liu, Wang, Li, Wu, Zhang, Duan, Zhang, Liu, Jiang, Yan, Zhang, Wei, Chen, Feng, Yao, Chai, Wang, Zhang, Xu, Huang, Wang, Li, Dong, and Tang]{glm}
GLM-5-Team, :, Aohan Zeng, Xin Lv, Zhenyu Hou, Zhengxiao Du, Qinkai Zheng, Bin Chen, Da~Yin, Chendi Ge, Chenghua Huang, Chengxing Xie, Chenzheng Zhu, Congfeng Yin, Cunxiang Wang, Gengzheng Pan, Hao Zeng, Haoke Zhang, Haoran Wang, Huilong Chen, Jiajie Zhang, Jian Jiao, Jiaqi Guo, Jingsen Wang, Jingzhao Du, Jinzhu Wu, Kedong Wang, Lei Li, Lin Fan, Lucen Zhong, Mingdao Liu, Mingming Zhao, Pengfan Du, Qian Dong, Rui Lu, Shuang-Li, Shulin Cao, Song Liu, Ting Jiang, Xiaodong Chen, Xiaohan Zhang, Xuancheng Huang, Xuezhen Dong, Yabo Xu, Yao Wei, Yifan An, Yilin Niu, Yitong Zhu, Yuanhao Wen, Yukuo Cen, Yushi Bai, Zhongpei Qiao, Zihan Wang, Zikang Wang, Zilin Zhu, Ziqiang Liu, Zixuan Li, Bojie Wang, Bosi Wen, Can Huang, Changpeng Cai, Chao Yu, Chen Li, Chengwei Hu, Chenhui Zhang, Dan Zhang, Daoyan Lin, Dayong Yang, Di~Wang, Ding Ai, Erle Zhu, Fangzhou Yi, Feiyu Chen, Guohong Wen, Hailong Sun, Haisha Zhao, Haiyi Hu, Hanchen Zhang, Hanrui Liu, Hanyu Zhang, Hao Peng, Hao Tai, Haobo Zhang, He~Liu, Hongwei Wang, Hongxi Yan, Hongyu Ge, Huan Liu, Huanpeng Chu, Jia'ni Zhao, Jiachen Wang, Jiajing Zhao, Jiamin Ren, Jiapeng Wang, Jiaxin Zhang, Jiayi Gui, Jiayue Zhao, Jijie Li, Jing An, Jing Li, Jingwei Yuan, Jinhua Du, Jinxin Liu, Junkai Zhi, Junwen Duan, Kaiyue Zhou, Kangjian Wei, Ke~Wang, Keyun Luo, Laiqiang Zhang, Leigang Sha, Liang Xu, Lindong Wu, Lintao Ding, Lu~Chen, Minghao Li, Nianyi Lin, Pan Ta, Qiang Zou, Rongjun Song, Ruiqi Yang, Shangqing Tu, Shangtong Yang, Shaoxiang Wu, Shengyan Zhang, Shijie Li, Shuang Li, Shuyi Fan, Wei Qin, Wei Tian, Weining Zhang, Wenbo Yu, Wenjie Liang, Xiang Kuang, Xiangmeng Cheng, Xiangyang Li, Xiaoquan Yan, Xiaowei Hu, Xiaoying Ling, Xing Fan, Xingye Xia, Xinyuan Zhang, Xinze Zhang, Xirui Pan, Xu~Zou, Xunkai Zhang, Yadi Liu, Yandong Wu, Yanfu Li, Yidong Wang, Yifan Zhu, Yijun Tan, Yilin Zhou, Yiming Pan, Ying Zhang, Yinpei Su, Yipeng Geng, Yong Yan, Yonglin Tan, Yuean Bi, Yuhan Shen, Yuhao Yang, Yujiang Li, Yunan Liu, Yunqing Wang, Yuntao Li, Yurong Wu, Yutao Zhang, Yuxi Duan, Yuxuan Zhang, Zezhen Liu, Zhengtao Jiang, Zhenhe Yan, Zheyu Zhang, Zhixiang Wei, Zhuo Chen, Zhuoer Feng, Zijun Yao, Ziwei Chai, Ziyuan Wang, Zuzhou Zhang, Bin Xu, Minlie Huang, Hongning Wang, Juanzi Li, Yuxiao Dong, and Jie Tang.
\newblock Glm-5: from vibe coding to agentic engineering, 2026.
\newblock URL \url{https://arxiv.org/abs/2602.15763}.

\bibitem[Google(2026)]{gemini}
Google.
\newblock Gemini 3.7 flash model card, 2026.
\newblock URL \url{https://storage.googleapis.com/deepmind-media/Model-Cards/Gemini-3-7-Flash-Model-Card.pdf}.

\bibitem[Guo et~al.(2024)Guo, He, Zhang, Zhang, Feng, Huang, and Ma]{detective}
Xun Guo, Yongxin He, Shan Zhang, Ting Zhang, Wanquan Feng, Haibin Huang, and Chongyang Ma.
\newblock Detective: Detecting {AI}-generated text via multi-level contrastive learning.
\newblock In \emph{The Thirty-eighth Annual Conference on Neural Information Processing Systems}, 2024.
\newblock URL \url{https://openreview.net/forum?id=cdTTTJfJe3}.

\bibitem[Hans et~al.(2024)Hans, Schwarzschild, Cherepanova, Kazemi, Saha, Goldblum, Geiping, and Goldstein]{binoculars}
Abhimanyu Hans, Avi Schwarzschild, Valeriia Cherepanova, Hamid Kazemi, Aniruddha Saha, Micah Goldblum, Jonas Geiping, and Tom Goldstein.
\newblock Spotting {LLM}s with binoculars: Zero-shot detection of machine-generated text.
\newblock In \emph{Forty-first International Conference on Machine Learning}, 2024.
\newblock URL \url{https://openreview.net/forum?id=axl3FAkpik}.

\bibitem[He et~al.(2024)He, Shen, Chen, Backes, and Zhang]{mgtbench}
Xinlei He, Xinyue Shen, Zeyuan Chen, Michael Backes, and Yang Zhang.
\newblock Mgtbench: Benchmarking machine-generated text detection, 2024.
\newblock URL \url{https://arxiv.org/abs/2303.14822}.

\bibitem[Hou et~al.(2026)Hou, Li, Fu, He, Yan, Chen, and McAuley]{review}
Yupeng Hou, Jiacheng Li, Xiangjun Fu, Zhankui He, An~Yan, Xiusi Chen, and Julian McAuley.
\newblock Bridging language and items for retrieval and recommendation: Benchmarking {LLM}s as semantic encoders.
\newblock In Maria Liakata, Viviane~P. Moreira, Jiajun Zhang, and David Jurgens (eds.), \emph{Proceedings of the 64th Annual Meeting of the {A}ssociation for {C}omputational {L}inguistics (Volume 1: Long Papers)}, pp.\  3251--3265, San Diego, California, United States, July 2026. Association for Computational Linguistics.
\newblock ISBN 979-8-89176-390-6.
\newblock \doi{10.18653/v1/2026.acl-long.147}.
\newblock URL \url{https://aclanthology.org/2026.acl-long.147/}.

\bibitem[Hu et~al.(2022)Hu, yelong shen, Wallis, Allen-Zhu, Li, Wang, Wang, and Chen]{lora}
Edward~J Hu, yelong shen, Phillip Wallis, Zeyuan Allen-Zhu, Yuanzhi Li, Shean Wang, Lu~Wang, and Weizhu Chen.
\newblock Lo{RA}: Low-rank adaptation of large language models.
\newblock In \emph{International Conference on Learning Representations}, 2022.
\newblock URL \url{https://openreview.net/forum?id=nZeVKeeFYf9}.

\bibitem[Jiang et~al.(2023)Jiang, Sablayrolles, Mensch, Bamford, Chaplot, de~las Casas, Bressand, Lengyel, Lample, Saulnier, Lavaud, Lachaux, Stock, Scao, Lavril, Wang, Lacroix, and Sayed]{mistral}
Albert~Q. Jiang, Alexandre Sablayrolles, Arthur Mensch, Chris Bamford, Devendra~Singh Chaplot, Diego de~las Casas, Florian Bressand, Gianna Lengyel, Guillaume Lample, Lucile Saulnier, Lélio~Renard Lavaud, Marie-Anne Lachaux, Pierre Stock, Teven~Le Scao, Thibaut Lavril, Thomas Wang, Timothée Lacroix, and William~El Sayed.
\newblock Mistral 7b, 2023.
\newblock URL \url{https://arxiv.org/abs/2310.06825}.

\bibitem[Kirchenbauer et~al.(2023)Kirchenbauer, Geiping, Wen, Katz, Miers, and Goldstein]{watermark}
John Kirchenbauer, Jonas Geiping, Yuxin Wen, Jonathan Katz, Ian Miers, and Tom Goldstein.
\newblock A watermark for large language models.
\newblock In Andreas Krause, Emma Brunskill, Kyunghyun Cho, Barbara Engelhardt, Sivan Sabato, and Jonathan Scarlett (eds.), \emph{Proceedings of the 40th International Conference on Machine Learning}, volume 202 of \emph{Proceedings of Machine Learning Research}, pp.\  17061--17084. PMLR, 23--29 Jul 2023.
\newblock URL \url{https://proceedings.mlr.press/v202/kirchenbauer23a.html}.

\bibitem[La~Cava \& Tagarelli(2025)La~Cava and Tagarelli]{openturingbench}
Lucio La~Cava and Andrea Tagarelli.
\newblock {O}pen{T}uring{B}ench: An open-model-based benchmark and framework for machine-generated text detection and attribution.
\newblock In Christos Christodoulopoulos, Tanmoy Chakraborty, Carolyn Rose, and Violet Peng (eds.), \emph{Proceedings of the 2025 Conference on Empirical Methods in Natural Language Processing}, pp.\  26655--26671, Suzhou, China, November 2025. Association for Computational Linguistics.
\newblock ISBN 979-8-89176-332-6.
\newblock \doi{10.18653/v1/2025.emnlp-main.1354}.
\newblock URL \url{https://aclanthology.org/2025.emnlp-main.1354/}.

\bibitem[La~Cava et~al.(2026)La~Cava, Macko, Moro, Srba, and Tagarelli]{multilingual}
Lucio La~Cava, Dominik Macko, Robert Moro, Ivan Srba, and Andrea Tagarelli.
\newblock Authorship attribution in multilingual machine-generated texts.
\newblock In Maria Liakata, Viviane~P. Moreira, Jiajun Zhang, and David Jurgens (eds.), \emph{Proceedings of the 64th Annual Meeting of the {A}ssociation for {C}omputational {L}inguistics (Volume 1: Long Papers)}, pp.\  45136--45152, San Diego, California, United States, July 2026. Association for Computational Linguistics.
\newblock ISBN 979-8-89176-390-6.
\newblock \doi{10.18653/v1/2026.acl-long.2091}.
\newblock URL \url{https://aclanthology.org/2026.acl-long.2091/}.

\bibitem[Lee et~al.(2024)Lee, Tack, and Shin]{remodetect}
Hyunseok Lee, Jihoon Tack, and Jinwoo Shin.
\newblock Remodetect: Reward models recognize aligned {LLM}'s generations.
\newblock In \emph{The Thirty-eighth Annual Conference on Neural Information Processing Systems}, 2024.
\newblock URL \url{https://openreview.net/forum?id=pW9Jwim918}.

\bibitem[Li et~al.(2025)Li, Kumar, and Pathak]{generative}
Alexander~Cong Li, Ananya Kumar, and Deepak Pathak.
\newblock Generative classifiers avoid shortcut solutions.
\newblock In \emph{The Thirteenth International Conference on Learning Representations}, 2025.
\newblock URL \url{https://openreview.net/forum?id=oCUYc7BzXQ}.

\bibitem[Li et~al.(2023)Li, Wang, Ren, Sun, and Qiu]{sniffer}
Linyang Li, Pengyu Wang, Ke~Ren, Tianxiang Sun, and Xipeng Qiu.
\newblock Origin tracing and detecting of llms, 2023.
\newblock URL \url{https://arxiv.org/abs/2304.14072}.

\bibitem[Lim et~al.(2026)Lim, Chung, Lee, Cheung, Cho, Ha, Kang, Kim, Kim, Kim, Kim, Kim, Lee, Lee, Lee, Oh, Cho, Choi, Her, Jung, Kang, Kim, Kim, Kweon, Lee, Park, and Ryu]{motif3}
Junghwan Lim, Joon~Son Chung, Sungmin Lee, Wai~Ting Cheung, Gihun Cho, Minsu Ha, Sangho Kang, Beomgyu Kim, Dongseok Kim, Jangwoong Kim, Taehyun Kim, Taewhan Kim, Jeesoo Lee, Jeongdoo Lee, Junhyeok Lee, Dongpin Oh, Hyeyeon Cho, Dahye Choi, Jaeheui Her, Hanbin Jung, Changjin Kang, Minjae Kim, Youngrok Kim, Hyukjin Kweon, Hongjoo Lee, Yeongjae Park, and Bokki Ryu.
\newblock {Motif 3}: Technical report, 2026.
\newblock URL \url{https://arxiv.org/abs/2608.09119}.

\bibitem[Liu et~al.(2025)Liu, Huang, Xiao, and Wu]{irm}
Runheng Liu, Heyan Huang, Xingchen Xiao, and Zhijing Wu.
\newblock Zero-shot detection of {LLM}-generated text via implicit reward model.
\newblock In \emph{The Thirty-ninth Annual Conference on Neural Information Processing Systems}, 2025.
\newblock URL \url{https://openreview.net/forum?id=2VdsYVXLDl}.

\bibitem[Liu et~al.(2020)Liu, Ott, Goyal, Du, Joshi, Chen, Levy, Lewis, Zettlemoyer, and Stoyanov]{roberta}
Yinhan Liu, Myle Ott, Naman Goyal, Jingfei Du, Mandar Joshi, Danqi Chen, Omer Levy, Mike Lewis, Luke Zettlemoyer, and Veselin Stoyanov.
\newblock Ro{\{}bert{\}}a: A robustly optimized {\{}bert{\}} pretraining approach, 2020.
\newblock URL \url{https://openreview.net/forum?id=SyxS0T4tvS}.

\bibitem[{Meta}(2025)]{llama}
{Meta}.
\newblock The llama 4 herd: The beginning of a new era of natively multimodal ai innovation, 2025.
\newblock URL \url{https://ai.meta.com/blog/llama-4-multimodal-intelligence/}.

\bibitem[Mitchell et~al.(2023)Mitchell, Lee, Khazatsky, Manning, and Finn]{detectgpt}
Eric Mitchell, Yoonho Lee, Alexander Khazatsky, Christopher~D Manning, and Chelsea Finn.
\newblock {D}etect{GPT}: Zero-shot machine-generated text detection using probability curvature.
\newblock In Andreas Krause, Emma Brunskill, Kyunghyun Cho, Barbara Engelhardt, Sivan Sabato, and Jonathan Scarlett (eds.), \emph{Proceedings of the 40th International Conference on Machine Learning}, volume 202 of \emph{Proceedings of Machine Learning Research}, pp.\  24950--24962. PMLR, 23--29 Jul 2023.
\newblock URL \url{https://proceedings.mlr.press/v202/mitchell23a.html}.

\bibitem[Narayan et~al.(2018)Narayan, Cohen, and Lapata]{xsum}
Shashi Narayan, Shay~B. Cohen, and Mirella Lapata.
\newblock Don't give me the details, just the summary! topic-aware convolutional neural networks for extreme summarization, 2018.
\newblock URL \url{https://arxiv.org/abs/1808.08745}.

\bibitem[OpenAI(2025)]{gpt5}
OpenAI.
\newblock Gpt-5 system card, 2025.
\newblock URL \url{https://cdn.openai.com/gpt-5-system-card.pdf}.

\bibitem[Qwen et~al.(2025)Qwen, :, Yang, Yang, Zhang, Hui, Zheng, Yu, Li, Liu, Huang, Wei, Lin, Yang, Tu, Zhang, Yang, Yang, Zhou, Lin, Dang, Lu, Bao, Yang, Yu, Li, Xue, Zhang, Zhu, Men, Lin, Li, Tang, Xia, Ren, Ren, Fan, Su, Zhang, Wan, Liu, Cui, Zhang, and Qiu]{qwen2.5}
Qwen, :, An~Yang, Baosong Yang, Beichen Zhang, Binyuan Hui, Bo~Zheng, Bowen Yu, Chengyuan Li, Dayiheng Liu, Fei Huang, Haoran Wei, Huan Lin, Jian Yang, Jianhong Tu, Jianwei Zhang, Jianxin Yang, Jiaxi Yang, Jingren Zhou, Junyang Lin, Kai Dang, Keming Lu, Keqin Bao, Kexin Yang, Le~Yu, Mei Li, Mingfeng Xue, Pei Zhang, Qin Zhu, Rui Men, Runji Lin, Tianhao Li, Tianyi Tang, Tingyu Xia, Xingzhang Ren, Xuancheng Ren, Yang Fan, Yang Su, Yichang Zhang, Yu~Wan, Yuqiong Liu, Zeyu Cui, Zhenru Zhang, and Zihan Qiu.
\newblock Qwen2.5 technical report, 2025.
\newblock URL \url{https://arxiv.org/abs/2412.15115}.

\bibitem[Rao et~al.(2025)Rao, Mohamed, Liu, and Liu]{damtl}
Zixin Rao, Youssef Mohamed, Shang Liu, and Zeyan Liu.
\newblock Two birds with one stone: Multi-task detection and attribution of llm-generated text, 2025.
\newblock URL \url{https://arxiv.org/abs/2508.14190}.

\bibitem[Sajadieh et~al.(2026)Sajadieh, Fattorini, Perrault, Gil, Parli, Santarlasci, Pava, Maslej, Altman, Brynjolfsson, Brodley, Clark, Dignum, Kumar, Landay, Lyons, Manyika, Niebles, Shoham, Tabassi, Wald, Walsh, and Weld]{aiindex}
Sha Sajadieh, Loredana Fattorini, Raymond Perrault, Yolanda Gil, Vanessa Parli, Lapo Santarlasci, Juan Pava, Nestor Maslej, Russ Altman, Erik Brynjolfsson, Carla Brodley, Jack Clark, Virginia Dignum, Vipin Kumar, James Landay, Terah Lyons, James Manyika, Juan~Carlos Niebles, Yoav Shoham, Elham Tabassi, Russell Wald, Toby Walsh, and Dan Weld.
\newblock Artificial intelligence index report 2026, 2026.
\newblock URL \url{https://arxiv.org/abs/2606.15708}.

\bibitem[{Stack Exchange, Inc.}(2014)]{stackexchange}
{Stack Exchange, Inc.}
\newblock Stack exchange data dump, 2014.
\newblock URL \url{https://archive.org/details/stackexchange}.

\bibitem[Su et~al.(2023)Su, Zhuo, Wang, and Nakov]{detectllm}
Jinyan Su, Terry~Yue Zhuo, Di~Wang, and Preslav Nakov.
\newblock Detect{LLM}: Leveraging log rank information for zero-shot detection of machine-generated text.
\newblock In \emph{The 2023 Conference on Empirical Methods in Natural Language Processing}, 2023.
\newblock URL \url{https://openreview.net/forum?id=Dy2mbQIdMz}.

\bibitem[Sun et~al.(2026{\natexlab{a}})Sun, Bao, Cui, and Zhang]{minimizing}
Ke~Sun, Guangsheng Bao, Han Cui, and Yue Zhang.
\newblock Minimizing mismatch risk: A prototype-based routing framework for zero-shot {LLM}-generated text detection.
\newblock In \emph{Forty-third International Conference on Machine Learning}, 2026{\natexlab{a}}.
\newblock URL \url{https://openreview.net/forum?id=Hh1hf4kauu}.

\bibitem[Sun et~al.(2025)Sun, Yin, Xu, Kolter, and Liu]{idiosyncrasies}
Mingjie Sun, Yida Yin, Zhiqiu Xu, J~Zico Kolter, and Zhuang Liu.
\newblock Idiosyncrasies in large language models.
\newblock In \emph{Forty-second International Conference on Machine Learning}, 2025.
\newblock URL \url{https://openreview.net/forum?id=FCZ3jVzmTZ}.

\bibitem[Sun et~al.(2026{\natexlab{b}})Sun, Liao, Huang, Wei, Hong, Yue, and He]{ridgeft}
Zhen Sun, Yifan Liao, Zhicong Huang, Jiaheng Wei, Cheng Hong, Yutao Yue, and Xinlei He.
\newblock When new generators arrive: Lifelong machine-generated text attribution via ridge feature transfer, 2026{\natexlab{b}}.
\newblock URL \url{https://arxiv.org/abs/2606.05626}.

\bibitem[Suzgun et~al.(2023)Suzgun, Melas-Kyriazi, Sarkar, Kominers, and Shieber]{hupd}
Mirac Suzgun, Luke Melas-Kyriazi, Suproteem~K Sarkar, Scott Kominers, and Stuart Shieber.
\newblock The harvard {USPTO} patent dataset: A large-scale, well-structured, and multi-purpose corpus of patent applications.
\newblock In \emph{Thirty-seventh Conference on Neural Information Processing Systems Datasets and Benchmarks Track}, 2023.
\newblock URL \url{https://openreview.net/forum?id=tk27oD2cBw}.

\bibitem[Team et~al.(2026)Team, Bai, Bao, Charles, Chen, Chen, Chen, Chen, Chen, Chen, Chen, Chen, Chen, Chen, Chen, Cui, Ding, Dong, Du, Du, Du, Du, Fan, Feng, Fu, Gao, Gao, Gao, Gao, Gao, Ge, Geng, Gu, Gu, Guan, Guo, Guo, Hao, He, He, He, He, Hong, Hu, Hu, Hu, Huang, Huang, Huang, Jiang, Jiang, Jin, Kang, Lai, Li, Li, Li, Li, Li, Li, Li, Li, Li, Li, Lin, Lin, Lin, Liu, Liu, Liu, Liu, Liu, Liu, Liu, Liu, Liu, Liu, Liu, Liu, Liu, Liu, Liu, Lu, Lu, Lu, Luo, Ma, Ma, Ma, Mao, Mei, Men, Miao, Pan, Peng, Qin, Qin, Qu, Shang, Shi, Shi, Song, Su, Su, Sui, Sun, Sung, Tai, Tang, Tao, Teng, Tian, Wang, Wang, Wang, Wang, Wang, Wang, Wang, Wang, Wang, Wang, Wang, Wang, Wang, Wang, Wang, Wang, Wang, Wang, Wang, Wang, Wang, Wei, Wei, Wu, Wu, Wu, Wu, Xiao, Xie, Xie, Xiong, Xu, Xu, Xu, Xu, Xu, Xu, Xu, Xu, Xu, Xu, Xu, Yan, Yan, Yang, Yang, Yang, Yang, Yang, Yang, Yang, Yao, Yao, Ye, Ye, Yin, Yu, Yuan, Yuan, Yuan, Yuan, Zhan, Zhang, Zhang, Zhang, Zhang, Zhang, Zhang, Zhang, Zhang, Zhang, Zhang, Zhang, Zhang, Zhang, Zhao, Zhao, Zhao, Zheng, Zheng, Zhong, Zhou, Zhou, Zhou, Zhu, Zhu, Zhuang, and Zu]{kimi}
Kimi Team, Yifan Bai, Yiping Bao, Y.~Charles, Cheng Chen, Guanduo Chen, Haiting Chen, Huarong Chen, Jiahao Chen, Ningxin Chen, Ruijue Chen, Yanru Chen, Yuankun Chen, Yutian Chen, Zhuofu Chen, Jialei Cui, Hao Ding, Mengnan Dong, Angang Du, Chenzhuang Du, Dikang Du, Yulun Du, Yu~Fan, Yichen Feng, Kelin Fu, Bofei Gao, Chenxiao Gao, Hongcheng Gao, Peizhong Gao, Tong Gao, Yuyao Ge, Shangyi Geng, Qizheng Gu, Xinran Gu, Longyu Guan, Haiqing Guo, Jianhang Guo, Xiaoru Hao, Tianhong He, Weiran He, Wenyang He, Yunjia He, Chao Hong, Hao Hu, Yangyang Hu, Zhenxing Hu, Weixiao Huang, Zhiqi Huang, Zihao Huang, Tao Jiang, Zhejun Jiang, Xinyi Jin, Yongsheng Kang, Guokun Lai, Cheng Li, Fang Li, Haoyang Li, Ming Li, Wentao Li, Yang Li, Yanhao Li, Yiwei Li, Zhaowei Li, Zheming Li, Hongzhan Lin, Xiaohan Lin, Zongyu Lin, Chengyin Liu, Chenyu Liu, Hongzhang Liu, Jingyuan Liu, Junqi Liu, Liang Liu, Shaowei Liu, T.~Y. Liu, Tianwei Liu, Weizhou Liu, Yangyang Liu, Yibo Liu, Yiping Liu, Yue Liu, Zhengying Liu, Enzhe Lu, Haoyu Lu, Lijun Lu, Yashuo Luo, Shengling Ma, Xinyu Ma, Yingwei Ma, Shaoguang Mao, Jie Mei, Xin Men, Yibo Miao, Siyuan Pan, Yebo Peng, Ruoyu Qin, Zeyu Qin, Bowen Qu, Zeyu Shang, Lidong Shi, Shengyuan Shi, Feifan Song, Jianlin Su, Zhengyuan Su, Lin Sui, Xinjie Sun, Flood Sung, Yunpeng Tai, Heyi Tang, Jiawen Tao, Qifeng Teng, Chaoran Tian, Chensi Wang, Dinglu Wang, Feng Wang, Hailong Wang, Haiming Wang, Jianzhou Wang, Jiaxing Wang, Jinhong Wang, Shengjie Wang, Shuyi Wang, Si~Wang, Xinyuan Wang, Yao Wang, Yejie Wang, Yiqin Wang, Yuxin Wang, Yuzhi Wang, Zhaoji Wang, Zhengtao Wang, Zhengtao Wang, Zhexu Wang, Chu Wei, Qianqian Wei, Haoning Wu, Wenhao Wu, Xingzhe Wu, Yuxin Wu, Chenjun Xiao, Jin Xie, Xiaotong Xie, Weimin Xiong, Boyu Xu, Jinjing Xu, L.~H. Xu, Lin Xu, Suting Xu, Weixin Xu, Xinran Xu, Yangchuan Xu, Ziyao Xu, Jing Xu, Jing Xu, Junjie Yan, Yuzi Yan, Hao Yang, Xiaofei Yang, Yi~Yang, Ying Yang, Zhen Yang, Zhilin Yang, Zonghan Yang, Haotian Yao, Xingcheng Yao, Wenjie Ye, Zhuorui Ye, Bohong Yin, Longhui Yu, Enming Yuan, Hongbang Yuan, Mengjie Yuan, Siyu Yuan, Haobing Zhan, Dehao Zhang, Hao Zhang, Wanlu Zhang, Xiaobin Zhang, Yadong Zhang, Yangkun Zhang, Yichi Zhang, Yizhi Zhang, Yongting Zhang, Yu~Zhang, Yutao Zhang, Yutong Zhang, Zheng Zhang, Haotian Zhao, Yikai Zhao, Zijia Zhao, Huabin Zheng, Shaojie Zheng, Longguang Zhong, Jianren Zhou, Xinyu Zhou, Zaida Zhou, Jinguo Zhu, Zhen Zhu, Weiyu Zhuang, and Xinxing Zu.
\newblock Kimi k2: Open agentic intelligence, 2026.
\newblock URL \url{https://arxiv.org/abs/2507.20534}.

\bibitem[Tian et~al.(2024)Tian, Shi, Guo, Li, and zhong Xu]{hydralora}
Chunlin Tian, Zhan Shi, Zhijiang Guo, Li~Li, and Cheng zhong Xu.
\newblock Hydralo{RA}: An asymmetric lo{RA} architecture for efficient fine-tuning.
\newblock In \emph{The Thirty-eighth Annual Conference on Neural Information Processing Systems}, 2024.
\newblock URL \url{https://openreview.net/forum?id=qEpi8uWX3N}.

\bibitem[Uchendu et~al.(2021)Uchendu, Ma, Le, Zhang, and Lee]{turingbench}
Adaku Uchendu, Zeyu Ma, Thai Le, Rui Zhang, and Dongwon Lee.
\newblock {TURINGBENCH}: A benchmark environment for {T}uring test in the age of neural text generation.
\newblock In Marie-Francine Moens, Xuanjing Huang, Lucia Specia, and Scott Wen-tau Yih (eds.), \emph{Findings of the Association for Computational Linguistics: EMNLP 2021}, pp.\  2001--2016, Punta Cana, Dominican Republic, November 2021. Association for Computational Linguistics.
\newblock \doi{10.18653/v1/2021.findings-emnlp.172}.
\newblock URL \url{https://aclanthology.org/2021.findings-emnlp.172/}.

\bibitem[{Upstage Team}(2026)]{solar}
{Upstage Team}.
\newblock {Solar Open 2}: {Korea's} sovereign foundation model, built for agentic use, 2026.
\newblock URL \url{https://www.upstage.ai/blog/en/solar-open-2}.

\bibitem[Venkatraman et~al.(2024)Venkatraman, Uchendu, and Lee]{gpt-who}
Saranya Venkatraman, Adaku Uchendu, and Dongwon Lee.
\newblock {GPT}-who: An information density-based machine-generated text detector.
\newblock In Kevin Duh, Helena Gomez, and Steven Bethard (eds.), \emph{Findings of the Association for Computational Linguistics: NAACL 2024}, pp.\  103--115, Mexico City, Mexico, June 2024. Association for Computational Linguistics.
\newblock \doi{10.18653/v1/2024.findings-naacl.8}.
\newblock URL \url{https://aclanthology.org/2024.findings-naacl.8/}.

\bibitem[Wang et~al.(2024)Wang, Mansurov, Ivanov, Su, Shelmanov, Tsvigun, Whitehouse, Mohammed~Afzal, Mahmoud, Sasaki, Arnold, Aji, Habash, Gurevych, and Nakov]{m4}
Yuxia Wang, Jonibek Mansurov, Petar Ivanov, Jinyan Su, Artem Shelmanov, Akim Tsvigun, Chenxi Whitehouse, Osama Mohammed~Afzal, Tarek Mahmoud, Toru Sasaki, Thomas Arnold, Alham Aji, Nizar Habash, Iryna Gurevych, and Preslav Nakov.
\newblock M4: Multi-generator, multi-domain, and multi-lingual black-box machine-generated text detection.
\newblock In Yvette Graham and Matthew Purver (eds.), \emph{Proceedings of the 18th Conference of the European Chapter of the Association for Computational Linguistics (Volume 1: Long Papers)}, pp.\  1369--1407, St. Julian{'}s, Malta, March 2024. Association for Computational Linguistics.
\newblock URL \url{https://aclanthology.org/2024.eacl-long.83}.

\bibitem[{Wikimedia Foundation}(2023)]{wikidump}
{Wikimedia Foundation}.
\newblock Wikimedia downloads, 2023.
\newblock URL \url{https://dumps.wikimedia.org}.

\bibitem[Wu et~al.(2024{\natexlab{a}})Wu, Yang, Zhan, Yuan, Wong, and Chao]{survey}
Junchao Wu, Shu Yang, Runzhe Zhan, Yulin Yuan, Derek~F. Wong, and Lidia~S. Chao.
\newblock A survey on llm-generated text detection: Necessity, methods, and future directions, 2024{\natexlab{a}}.
\newblock URL \url{https://arxiv.org/abs/2310.14724}.

\bibitem[Wu et~al.(2024{\natexlab{b}})Wu, Zhan, Wong, Yang, Yang, Yuan, and Chao]{detectrl}
Junchao Wu, Runzhe Zhan, Derek~F. Wong, Shu Yang, Xinyi Yang, Yulin Yuan, and Lidia~S. Chao.
\newblock Detect{RL}: Benchmarking {LLM}-generated text detection in real-world scenarios.
\newblock In \emph{The Thirty-eight Conference on Neural Information Processing Systems Datasets and Benchmarks Track}, 2024{\natexlab{b}}.
\newblock URL \url{https://openreview.net/forum?id=ZGMkOikEyv}.

\bibitem[Wu et~al.(2023)Wu, Pang, Shen, Cheng, and Chua]{llmdet}
Kangxi Wu, Liang Pang, Huawei Shen, Xueqi Cheng, and Tat-Seng Chua.
\newblock {LLMD}et: A third party large language models generated text detection tool.
\newblock In Houda Bouamor, Juan Pino, and Kalika Bali (eds.), \emph{Findings of the Association for Computational Linguistics: EMNLP 2023}, pp.\  2113--2133, Singapore, December 2023. Association for Computational Linguistics.
\newblock \doi{10.18653/v1/2023.findings-emnlp.139}.
\newblock URL \url{https://aclanthology.org/2023.findings-emnlp.139/}.

\bibitem[{xAI}(2026)]{grok}
{xAI}.
\newblock Model card: {Grok 4.7}, 2026.
\newblock URL \url{https://media.x.ai/v1/website/4p7card-5eccc980.pdf}.

\bibitem[Yang et~al.(2024)Yang, Cheng, Wu, Petzold, Wang, and Chen]{dnagpt}
Xianjun Yang, Wei Cheng, Yue Wu, Linda~Ruth Petzold, William~Yang Wang, and Haifeng Chen.
\newblock {DNA}-{GPT}: Divergent n-gram analysis for training-free detection of {GPT}-generated text.
\newblock In \emph{The Twelfth International Conference on Learning Representations}, 2024.
\newblock URL \url{https://openreview.net/forum?id=Xlayxj2fWp}.

\bibitem[Yu et~al.(2025)Yu, Yu, Liu, Chen, Zhang, Yu, and Shao]{evobench}
Xiao Yu, Yi~Yu, Dongrui Liu, Kejiang Chen, Weiming Zhang, Nenghai Yu, and Jing Shao.
\newblock {E}vo{B}ench: Towards real-world {LLM}-generated text detection benchmarking for evolving large language models.
\newblock In Wanxiang Che, Joyce Nabende, Ekaterina Shutova, and Mohammad~Taher Pilehvar (eds.), \emph{Findings of the Association for Computational Linguistics: ACL 2025}, pp.\  14605--14620, Vienna, Austria, July 2025. Association for Computational Linguistics.
\newblock ISBN 979-8-89176-256-5.
\newblock \doi{10.18653/v1/2025.findings-acl.754}.
\newblock URL \url{https://aclanthology.org/2025.findings-acl.754/}.

\bibitem[Zhu et~al.(2024)Zhu, Greenewald, Nadjahi, S\'{a}ez De Oc\'{a}riz~Borde, Gabrielsson, Choshen, Ghassemi, Yurochkin, and Solomon]{asymmetry}
Jiacheng Zhu, Kristjan Greenewald, Kimia Nadjahi, Haitz S\'{a}ez De Oc\'{a}riz~Borde, Rickard~Br\"{u}el Gabrielsson, Leshem Choshen, Marzyeh Ghassemi, Mikhail Yurochkin, and Justin Solomon.
\newblock Asymmetry in low-rank adapters of foundation models.
\newblock In Ruslan Salakhutdinov, Zico Kolter, Katherine Heller, Adrian Weller, Nuria Oliver, Jonathan Scarlett, and Felix Berkenkamp (eds.), \emph{Proceedings of the 41st International Conference on Machine Learning}, volume 235 of \emph{Proceedings of Machine Learning Research}, pp.\  62369--62385. PMLR, 21--27 Jul 2024.
\newblock URL \url{https://proceedings.mlr.press/v235/zhu24c.html}.

\end{thebibliography}

\appendix
\clearpage
\startcontents[appendix]
\section*{Appendix Contents}
\printcontents[appendix]{}{1}{\setcounter{tocdepth}{2}}
\clearpage
\section{\benchname{} Details}
\label{app:bench}

Appendices~\ref{sec:bench_source_models}--\ref{sec:bench_generation} detail the data construction of Section~\ref{sec:benchmark_data}, and Appendix~\ref{sec:bench_leakage} describes how the evaluation settings of Section~\ref{sec:benchmark_settings} are kept free of data leakage.

\subsection{Source Models}
\label{sec:bench_source_models}

We construct \benchname{} from five LLM families representing both black-box and openly accessible white-box models: GPT~\citep{gpt5}, Claude~\citep{claude}, Gemini~\citep{gemini}, Mistral~\citep{mistral}, and Llama~\citep{llama}.
For each family, we consider one model version as the training-time source model and evaluate under three subsequently released versions for the Version shift and Joint shift settings.

\begin{table}[ht]
\centering
\caption{\textbf{Source models of \benchname{}.} The training version is used for ID and Domain shift, and the three held-out versions for Version shift and Joint shift.}
\label{tab:bench_source_models}
\footnotesize
\setlength{\tabcolsep}{3pt}
\begin{tabular}{l l l l}
\toprule
\textbf{Family} & \textbf{Training Version} & \textbf{Version shift} & \textbf{Access} \\
\midrule
GPT
& GPT-5
& GPT-5.2 / GPT-5.4 / GPT-5.6-terra
& API \\
Claude
& Opus 4.6
& Opus 4.7 / Opus 4.8 / Opus 5
& API \\
Gemini
& 2.5 Flash
& 3 Flash / 3.5 Flash / 3.6 Flash
& API \\
Mistral
& Small-24B-2501
& Small-3.1-24B-2503 / Small-3.2-24B-2506 / small-2603
& Local \\
Llama
& Llama-3-70B
& Llama-3.1-70B / Llama-3.3-70B / Llama-4-Scout-17B
& Local \\
\bottomrule
\end{tabular}
\end{table}

Table~\ref{tab:bench_source_models} summarizes the models used in each family.
Black-box models are accessed through their respective APIs, while openly accessible white-box models are locally served.
The three held-out versions are generated independently of the training versions.

\subsection{Human-Written Text Corpus}
\label{sec:bench_human_corpus}
We construct \benchname{} from human-written texts that serve as seeds for generating the machine-generated texts. For example, a news seed consists of an XSum article and its headline. Every source model receives the same prompt constructed from the headline, while the original article is retained as the human-written text.

We consider six domains including general-purpose, technical, commercial, and community-generated text, each drawn from one public corpus: news from XSum~\citep{xsum}, Wikipedia from the English Wikipedia dump~\citep{wikidump}, academic writing from arXiv abstracts~\citep{arxiv}, product reviews from Amazon Reviews~\citep{review}, patents from HUPD~\citep{hupd}, and Stack Exchange from its public data dump~\citep{stackexchange}.
For each seed, we retain the original text together with domain-specific metadata used to construct the generation prompt.
Note that the human-written text itself is not provided to the source models, and only the extracted metadata are used as generation inputs.
The original human-written text is retained separately as the human class.

\begin{table}[t]
\centering
\caption{\textbf{Human-written text corpus of \benchname{}.} Patents and Stack Exchange are the unseen domains of Domain shift and Joint shift; counts are the seed pool before filtering.}
\label{tab:bench_human_corpus}
\footnotesize
\setlength{\tabcolsep}{5pt}
\begin{tabular}{l l l r}
\toprule
\textbf{Domain} & \textbf{Corpus} & \textbf{Generation Fields} & \textbf{\# Seeds} \\
\midrule
News            & XSum\footnotemark                    & Headline                  & 8,500 \\
Wikipedia       & Wikipedia EN\footnotemark & Title                     & 8,500 \\
Academic        & arXiv abstracts\footnotemark         & Title, Field              & 8,500 \\
Product Reviews & Amazon Reviews 2023\footnotemark     & Product, Category, Rating & 8,500 \\
Patents         & HUPD\footnotemark         & Title, Classification     &   750 \\
Stack Exchange  & Stack Exchange title/body dumps\footnotemark     & Title, Tags, Community    &   750 \\
\midrule
\textbf{Total}  &                                  &                           & \textbf{35,500} \\
\bottomrule
\end{tabular}
\end{table}

\addtocounter{footnote}{-6}
\stepcounter{footnote}\footnotetext{\texttt{EdinburghNLP/xsum@7d4d486c}}
\stepcounter{footnote}\footnotetext{\texttt{wikimedia/wikipedia:20231101.en@b04c8d1c}}
\stepcounter{footnote}\footnotetext{\texttt{gfissore/arxiv-abstracts-2021@e4c5fbd4}}
\stepcounter{footnote}\footnotetext{\texttt{McAuley-Lab/Amazon-Reviews-2023@2b6d039e}}
\stepcounter{footnote}\footnotetext{\texttt{HUPD/hupd@f570a84b}}
\stepcounter{footnote}\footnotetext{\texttt{flax-sentence-embeddings/stackexchange\_title\_body\_jsonl@a3d99bf2}}

We collect 8,500 seeds for each of the four ID domains and 750 seeds for each of the two unseen domains.
The seeds for the ID domain are divided into training, validation, and test splits at the seed level, ensuring that all generations derived from the same seed remain within the same split.
Human-written texts are restricted to 90--220 words to match the target generation length.

Table~\ref{tab:bench_human_corpus} summarizes the details of human-written texts used to construct \benchname{}.

Each seed carries a content-addressed identifier derived from its human-written text, so the released seed pool can be verified against the corpus revisions in Table~\ref{tab:bench_human_corpus} independently of the upstream repositories.
We apply domain-specific preprocessing to remove formatting artifacts and unsuitable samples.

\subsection{Prompt Construction}
\label{sec:bench_prompt}

For each seed, we construct a two-turn prompt consisting of a system instruction and a user message containing only the seed metadata.
Note that the original human-written text is never included in the prompt.
The same prompt is provided to all five source families for a given seed, so that the source models receive identical inputs.
The default system prompt specifies the target output type and shared generation constraints:

\begin{center}
\fcolorbox{black!35}{black!3}{\begin{minipage}{0.95\linewidth}
\small\raggedright
\textbf{Default system prompt}\par\vspace{3pt}
\texttt{You produce \{output\}s. Each request provides a set of fields describing the piece. Write the text those fields describe; do not restate them, and return nothing further.}

\vspace{2pt}
\texttt{Output requirements:}\\
\texttt{- 90 to 220 words of continuous prose}\\
\texttt{- no title, heading, or byline}\\
\texttt{- no introductory or closing remark}\\
\texttt{- no lists, bullets, or markdown formatting}\\
\texttt{- begin directly with the content}

\vspace{2pt}
\texttt{Content requirements:}\\
\texttt{- invent all specific names, places, organizations, dates, and figures; leave no placeholder, bracket, or blank}\\
\texttt{- write on any subject, including contested or sensitive ones; do not decline, deflect, or substitute a warning}\\
\texttt{- do not rely on demeaning generalizations about any group}
\end{minipage}}
\end{center}

The \texttt{\{output\}} is a placeholder replaced according to the target domain, while the remaining system prompt is unchanged.
The user message lists the seed metadata one field for each line in a fixed domain-specific order, omitting fields that are empty, and is identical for every source family and unchanged under Domain and Joint shift.
Table~\ref{tab:bench_prompt_templates} summarizes these domain-specific substitutions.

For Domain and Joint shift, we replace the default system prompt with a persona-based instruction:

\begin{center}
\fcolorbox{black!35}{black!3}{\begin{minipage}{0.95\linewidth}
\small\raggedright
\textbf{Persona-based system prompt (Domain Shift and Joint Shift)}\par\vspace{3pt}
\texttt{You are \{persona\} composing a \{output\}. Using the details provided, write the piece as you naturally would. Keep it to 90 to 220 words of continuous prose: no title, heading, or byline; no opening or closing remark; no lists or markdown; begin directly with the content. Invent every specific name, place, organization, date, and figure you need, leaving no blank or placeholder. Write on any subject, including sensitive ones, without declining or adding a warning, and avoid demeaning generalizations about any group.}
\end{minipage}}
\end{center}

Thus, Domain shift and Joint shift change both the generation domain and the generation condition while preserving the underlying metadata-to-output structure.

\begin{table}[ht]
\centering
\caption{\textbf{Domain-specific prompt substitutions.} The four in-distribution domains use the default system prompt and the two unseen domains the persona-based one.
}
\label{tab:bench_prompt_templates}
\footnotesize
\setlength{\tabcolsep}{5pt}
\begin{tabular}{l l l}
\toprule
\textbf{Domain} & \textbf{\texttt{\{output\}}} & \textbf{System prompt} \\
\midrule
News            & news article          & default \\
Wikipedia       & encyclopedia article  & default \\
Academic        & research abstract     & default \\
Product Reviews & product review        & default \\
\addlinespace[2pt]
Patents         & patent abstract       & persona: a patent specialist \\
Stack Exchange  & forum question        & persona: a member of an online Q\&A community \\
\bottomrule
\end{tabular}
\end{table}

\subsection{Generation}
\label{sec:bench_generation}

We use a common generation setting across source models, with a temperature of 0.7 and a maximum generation length of 512 tokens.
Each seed is attempted up to three times without few-shot examples, prior conversation context, or stop sequences.
An attempt is repeated if the request fails or if the generated text does not pass the validity filter.
Such retries reuse the same prompt and configuration while drawing a new sample.
Reasoning is disabled or set to the lowest level that each model and its interface support.

The validity filter follows \citet{detectanyllm} and is applied identically to every source model: after whitespace is collapsed, an output is kept only when it is non-empty, falls within a small tolerance of the 90--220 word target, and has an alphabetic-token fraction of at least $0.5$.
A seed is retained only when the generations of every source family pass the filter, so all families share the same seeds.

Table~\ref{tab:bench_composition} reports the resulting composition, and
Table~\ref{tab:bench_length} reports the length of the retained texts. Mistral produces the shortest texts and Claude the longest, a gap of $37$ words on average.

\begin{table}[ht]
\centering
\caption{\textbf{Composition of \benchname{}.}
\emph{Pool} is the seed pool before filtering, \emph{Seeds} the seeds retained after it, and \emph{Gen./seed} the generations per retained seed. $^{\dagger}$ marks a count reused from an earlier setting: Version shift reuses the in-distribution test seeds, and Joint shift is filtered from the same seed pool as Domain shift and shares all but two of its seeds.
}
\label{tab:bench_composition}
\footnotesize
\setlength{\tabcolsep}{4pt}
\begin{tabular}{l r r r r r r}
\toprule
\textbf{Setting} & \textbf{Pool} & \textbf{Seeds} & \textbf{Gen./seed} & \textbf{Machine} & \textbf{Human} & \textbf{Total} \\
\midrule
ID (train)      & 30,000 & 29,952 &  5 & 149,760 & 29,952 & 179,712 \\
ID (validation) &      2,000                    &  1,998 &  5 &   9,990 &  1,998 &  11,988 \\
ID (test)       &          2,000                &  1,997 &  5 &   9,985 &  1,997 &  11,982 \\
\addlinespace[2pt]
Domain shift    & 1,500 &  1,494 &  5 &   7,470 &  1,494 &   8,964 \\
Version shift   & 2,000 &  1,980 & 15 &  29,700 &  1,980$^{\dagger}$ &  31,680 \\
Joint shift     & 1,500 &  1,441 & 15 &  21,615 &  1,441$^{\dagger}$ &  23,056 \\
\midrule
\textbf{Total (unique)}  & & & & \textbf{228,520} & \textbf{35,443} & \textbf{263,963} \\
\bottomrule
\end{tabular}
\end{table}

\begin{table}[ht]
\centering
\caption{
\textbf{Length statistics} of the retained texts for each model.
}
\label{tab:bench_length}
\footnotesize
\setlength{\tabcolsep}{5pt}
\begin{tabular}{l r r r r r}
\toprule
\textbf{Source} & \textbf{$n$} & \textbf{Mean} & \textbf{Median} & \textbf{p5} & \textbf{p95} \\
\midrule
GPT     & 45,704 & 194.5 & 198 & 155 & 218 \\
Claude  & 45,704 & 198.8 & 199 & 180 & 218 \\
Gemini  & 45,704 & 192.5 & 192 & 162 & 226 \\
Mistral & 45,704 & 161.8 & 161 & 114 & 215 \\
Llama   & 45,704 & 175.9 & 173 & 147 & 214 \\
\addlinespace[2pt]
Human   & 35,443 & 166.9 & 180 &  96 & 217 \\
\bottomrule
\end{tabular}
\end{table}

\subsection{Data Leakage and Overlap Control}
\label{sec:bench_leakage}

The generation prompts contain only the seed metadata and the shared output requirements.
For each seed, source models receive the same prompt independently, without access to outputs from other models or previous generations.
We also exclude overlapping domain sources during corpus construction and apply temporal restrictions where applicable to reduce contamination of the human-written class.
These controls are intended to keep attribution performance reflecting source-model differences rather than explicit source identifiers or split-level seed overlap.

\clearpage
\section{Experimental Details}
\label{app:exp}

\subsection{Implementation Details and Evaluation Setup}
\label{app:exp_training}
This section supplements the implementation details and evaluation setup of Section~\ref{sec:exp_setup}.

\paragraph{Training.}
Among the parameters of a given pretrained LLM, we learn the down-projection $A$ of the asymmetric LoRA $\Delta W_k=B_kA$, which is shared across sources, together with the pair $c_k=(e_k,B_k).$
LoRA is applied to the attention projections ($q$, $k$, $v$, $o$) in every layer, with rank $16$, $\alpha=32$ and dropout $0.05$.
On the 7B backbone this leaves $24.79$M trainable parameters, $0.33\%$ of the model: $6.42$M in the shared $A$, $3.67$M in each of the five $B_k$, and $3{,}584$ in each control embedding.
The objective is $\mathcal L$ of Section~\ref{sec:method}.
Optimization is done with hyperparameters and settings in Table~\ref{tab:exp_optim} and runs for five epochs on \benchname{} and three on OpenTuringBench, on four H100-80GB GPUs.

\begin{table}[ht]
\centering
\caption{
\textbf{Optimization settings.} Shared by \mname{} and the supervised baselines.
}
\label{tab:exp_optim}
\footnotesize
\setlength{\tabcolsep}{5pt}
\begin{tabular}{l l}
\toprule
Optimizer & AdamW, $\beta=(0.9,0.999)$, weight decay $0.01$ \\
Learning rate & $1\times10^{-4}$ \\
Schedule & linear warmup over $5\%$ of steps, then cosine decay to $0$ \\
Gradient clipping & global norm $1.0$ \\
Effective batch & $128$ \\
Epochs & $5$ on \benchname{}, $3$ on OpenTuringBench \\
Maximum length & $512$ tokens (\benchname{}), $1024$ (OpenTuringBench) \\
Seed & $42$ ($42$, $123$, $456$ for Table~\ref{tab:main_results}) \\
\bottomrule
\end{tabular}
\end{table}

\paragraph{Inference.}
\label{app:method}
For each token $y_t$, let
$
\ell_{k,t}=\log p_\theta(y_t\mid y_{<t},c_k)
$
denote the token log-likelihood under source condition $c_k$, where $t$ ranges over the $n$ scored tokens of the text. For a temperature $T>0$, we compute the clipped score for source $k$ as
\[
r_{k,t}^{(T)}
=
\exp\left(
-\frac{1}{T}
\sum_{j\neq k}
[\ell_{j,t}-\ell_{k,t}]_+
\right),
\quad
\textup{where}
\quad
[z]_+=\max(z,0),
\]
which can be written equivalently as a product of pairwise likelihood ratios truncated at one as:
\[
r_{k,t}^{(T)}
=
\prod_{j\neq k}
\min\!\left(
\frac{p_\theta(y_t\mid y_{<t},c_k)}{p_\theta(y_t\mid y_{<t},c_j)},
\;1
\right)^{1/T}.
\]
The score equals one when source $k$ assigns the highest likelihood to $y_t$ and decreases as competing sources explain the token better. We then average the token-level scores over all scored tokens and temperature settings $\mathcal{T}=\{1,2,4,8\}$:
\[
R_k
=
\frac{1}{|\mathcal{T}|}
\sum_{T\in\mathcal{T}}
\frac{1}{n}
\sum_{t=1}^{n}r_{k,t}^{(T)}.
\]
The predicted source family is finally estimated by
$
\hat{k}=\arg\max_k R_k.
$

\clearpage
\subsection{Baselines}
\label{app:exp_baselines}
This section explains the baselines of Section~\ref{sec:exp_setup} with more details.
Table~\ref{tab:exp_baselines} summarizes them.

\begin{table}[ht]
\centering
\caption{\textbf{The MAA baselines.} Fixed-feature methods train only a classifier on features from frozen proxy models.
Encoder-tuning methods share the LLM2Vec trunk and the schedule of Table~\ref{tab:exp_optim}.}
\label{tab:exp_baselines}
\footnotesize
\setlength{\tabcolsep}{3pt}
\begin{tabular*}{\textwidth}{@{\extracolsep{\fill}}l c c c@{}}
\toprule
\textbf{Method} & \textbf{Representation} & \textbf{Trained part} & \textbf{Prediction} \\
\midrule
Fast-DetectGPT & one score from a frozen proxy    & classifier & logistic regression \\
Binoculars     & one score from two frozen proxies & classifier & logistic regression \\
GPT-who        & surprisal features, one proxy & classifier & logistic regression \\
Sniffer        & pairwise features, four proxies & classifier & MLP \\
StatEnsemble & eight features (Table~\ref{tab:exp_features}) & classifier & MLP \\
\addlinespace[2pt]
DeTeCtive      & LLM2Vec embedding & adapter and head & nearest neighbors \\
OTBDetector    & LLM2Vec embedding & adapter and head & nearest centroid \\
DA-MTL         & LLM2Vec embedding & adapter and heads & attribution head \\
Idiosyncrasies & LLM2Vec embedding & adapter and head & linear head \\
\bottomrule
\end{tabular*}
\end{table}

\paragraph{Fixed-feature baselines.}
Five of the baselines extract features from frozen proxy models. The standardization and the classifier are fitted on the in-distribution training split and reused unchanged on every shifted split. Fast-DetectGPT~\citep{fastdetectgpt} and Binoculars~\citep{binoculars} are zero-shot binary
detectors, each yielding a single scalar per text.
Following \citep{multilingual}, we train a logistic regression on that scalar to obtain a multiclass prediction.
GPT-who~\citep{gpt-who} and Sniffer~\citep{sniffer} instead produce multi-dimensional features and come with their own multiclass classifier, which we keep and widen to $K$ classes.
StatEnsemble~\citep{multilingual} concatenates a wider set of these statistics and classifies them with an MLP.

\begin{itemize}[leftmargin=*]
\item \textbf{Fast-DetectGPT}~\citep{fastdetectgpt}. The analytic curvature of the token log-likelihood, with \texttt{Qwen2.5-7B} as both the scoring and the sampling model. It is one scalar per text and we pass it to a multinomial logistic regression.

\item \textbf{Binoculars}~\citep{binoculars}. The ratio of perplexity to
cross-perplexity, with \texttt{Qwen2.5-7B} as the observer and \texttt{Qwen2.5-7B-Instruct} as the performer. This is again one scalar per text, passed to the same logistic regression under the same fit.

\item \textbf{GPT-who}~\citep{gpt-who}. Per-token surprisal under \texttt{Qwen2.5-7B}, reduced to four uniform information density statistics together with the raw surprisal of the lowest- and highest-variance token windows. The resulting features go to a logistic regression.

\item \textbf{Sniffer}~\citep{sniffer}. Per-word negative log-likelihoods from four proxies: \texttt{Qwen2.5-7B}, \texttt{Phi-3.5-mini-instruct}, \texttt{OLMo-2-1124-7B} and \texttt{Falcon3-7B-Base}. These are assembled into the pairwise representation of the original work and passed to a network with two hidden layers of width $300$, ReLU and dropout $0.5$. It is trained by Adam at $10^{-3}$ under cross-entropy in batches of $512$ for at most $200$ epochs, with early stopping at patience $20$ on a validation split.
\item \textbf{StatEnsemble}~\citep{multilingual}. The eight features of
Table~\ref{tab:exp_features}, computed from \texttt{Qwen2.5-7B} and \texttt{Qwen2.5-7B-Instruct}. They are passed to a multilayer perceptron with ReLU activations, trained by Adam at $10^{-3}$ for at most $1000$ iterations. Its architecture, one hidden layer of $100$ units or two of $100$ and $50$, and its $L_2$ coefficient, $10^{-4}$ or $10^{-3}$, are chosen by five-fold cross-validation.

\begin{table}[!ht]
\centering
\caption{\textbf{Features of StatEnsemble.} Every statistic is an average over the scored tokens of a text.}
\label{tab:exp_features}
\footnotesize
\setlength{\tabcolsep}{4pt}
\begin{tabular}{lc}
\toprule
\textbf{Feature} & \textbf{Definition} \\
\midrule
Log-likelihood & the mean log-probability of the written tokens \\
Perplexity     & the negative of the above, the mean token cross-entropy \\
Rank           & the mean rank of the written token, counting from one \\
Log-rank       & the mean of the logarithm of that rank, taken per token \\
Entropy        & the mean next-token entropy, which ignores the written token \\
LRR            & perplexity over log-rank, the statistic of DetectLLM~\citep{detectllm} \\
Fast-DetectGPT & analytic discrepancy, base model as both scoring and sampling model \\
Binoculars     & perplexity over cross-perplexity, base as observer and instruct as performer \\
\bottomrule
\end{tabular}
\end{table}

\end{itemize}
\paragraph{Encoder-tuning baselines.}
The other four learn a discriminative representation of a text and classify it.
They use encoders of different scale: Idiosyncrasies~\citep{idiosyncrasies} uses LLM2Vec \citep{llm2vec}, while the others are of the RoBERTa~\citep{roberta} scale.
Following Idiosyncrasies, we run all four on the same encoder, LLM2Vec over \texttt{Qwen2.5-7B-Instruct}, matching the backbone size of \mname{}.
We start from the released \texttt{mntp} checkpoint with its adapter merged,
mean-pool with left padding to a $3584$-dimensional embedding, and continue the released supervised adapter ($r=16$, $\alpha=32$, dropout $0.05$), whose rank matches \mname{}.
We train only its attention projections and follow Table~\ref{tab:exp_optim}.

\begin{itemize}[leftmargin=*]
\item \textbf{DeTeCtive}~\citep{detective}. Contrastive learning at temperature $0.07$ takes same-label texts as positives and the rest of the gathered $128$-text batch as negatives, with an auxiliary cross-entropy of equal weight through a three-layer head. That head is unused at inference: a text takes the majority label of its ten nearest indexed training embeddings under cosine similarity.

\item \textbf{OTBDetector}~\citep{openturingbench}. A batch-hard triplet loss with margin $1.0$ pulls same-source texts together and pushes different sources apart, over class-balanced batches of $120$ rather than the $128$ of Table~\ref{tab:exp_optim}. A text is then assigned to the nearest class centroid.

\item \textbf{DA-MTL}~\citep{damtl}. Two linear heads on the pooled embedding, one for MGT detection and one for MAA detection, whose cross-entropies are summed with equal weight over batches from two loaders. The MGT head is used only where human text is a class, and is disabled in the MAA experiments. Gradient surgery is off: it needs two backward passes per step and does not fit the distributed setting used here.
\item \textbf{Idiosyncrasies}~\citep{idiosyncrasies}. A single linear layer maps the pooled embedding to the $K$ source classes under cross-entropy. The same layer gives the prediction at inference.

\end{itemize}

\paragraph{MGT baselines.}
Two further detectors are used only for binary MGT detection, without training on \benchname{}.
\begin{itemize}[leftmargin=*]
\item \textbf{ReMoDetect}~\citep{remodetect}. The scalar head of an off-the-shelf reward model, \texttt{GRM-gemma2-2B-rewardmodel-ft}, is read directly as the detection score, with higher scores indicating machine-generated text. It relies on the observation that outputs of aligned LLMs consistently receive higher rewards than human-written text, since alignment training pushes a model toward high reward.

\item \textbf{IRM}~\citep{irm}. An implicit reward is computed as the difference in total log-likelihood between \texttt{Qwen2.5-7B-Instruct} and its base model \texttt{Qwen2.5-7B}, $\sum_t \log p_{\text{instruct}}(y_t \mid y_{<t}) - \sum_t \log p_{\text{base}}(y_t \mid y_{<t})$, which has the form of the DPO reward $\log \pi_\theta/\pi_{\text{ref}}$, and no separate reward model is needed.
\end{itemize}

\clearpage
\subsection{Results over Random Seeds}
\label{app:exp_seeds}

For \mname{} and the four encoder-tuning baselines, the entries of Table~\ref{tab:main_results} are the mean over three training runs with random seeds $42$, $123$, and $456$.
Table~\ref{tab:main_seeds} reports the result of each run together with the mean and standard deviation.
The standard deviation of \mname{} is at most $0.31$pp in every setting, and its margin over the strongest baseline on each OOD setting exceeds the run-to-run variation of every method.

\begin{table}[!ht]
\centering
\footnotesize
\setlength{\tabcolsep}{4.2pt}
\renewcommand{\arraystretch}{1.05}

\caption{
\textbf{Per-seed MAA performance on \benchname{}.}
Accuracy (\%, $\uparrow$) and macro-F1 (\%, $\uparrow$) of \mname{} and the four encoder-tuning baselines for each of the three random seeds, with their mean and standard deviation.
The means are the entries of Table~\ref{tab:main_results}.
}
\label{tab:main_seeds}

\begin{tabular}{
@{}l
@{\hspace{6pt}}
l
@{\hspace{8pt}}
cc
@{\hspace{5pt}}
cc
@{\hspace{5pt}}
cc
@{\hspace{5pt}}
cc@{}
}
\toprule

\multirow{3}{*}{Method}
& \multirow{3}{*}{Run}
& \multicolumn{2}{c}{\multirow{2}{*}{ID}}
& \multicolumn{6}{c}{OOD} \\
\cmidrule(lr){5-10}

& &
&
& \multicolumn{2}{c}{Domain}
& \multicolumn{2}{c}{Version}
& \multicolumn{2}{c}{Joint} \\

\cmidrule(lr){3-4}
\cmidrule(lr){5-6}
\cmidrule(lr){7-8}
\cmidrule(lr){9-10}

& & Acc & F1
& Acc & F1
& Acc & F1
& Acc & F1 \\

\midrule
\multirow{5}{*}{DeTeCtive} & Seed 42 & 99.49 & 99.49 & 95.39 & 95.40 & 70.12 & 69.46 & 62.87 & 61.54 \\
 & Seed 123 & 99.47 & 99.47 & 94.27 & 94.28 & 70.46 & 70.03 & 63.20 & 62.76 \\
 & Seed 456 & 99.56 & 99.56 & 94.42 & 94.47 & 71.22 & 71.14 & 63.89 & 63.60 \\
\cmidrule(lr){2-10}
 & Mean & 99.51 & 99.51 & 94.69 & 94.72 & 70.60 & 70.21 & 63.32 & 62.63 \\
 & Std & 0.05 & 0.05 & 0.61 & 0.60 & 0.56 & 0.86 & 0.52 & 1.04 \\
\midrule
\multirow{5}{*}{OTBDetector} & Seed 42 & 99.43 & 99.43 & 95.46 & 95.48 & 71.28 & 70.98 & 65.04 & 64.45 \\
 & Seed 123 & 99.56 & 99.56 & 95.01 & 95.00 & 68.75 & 67.88 & 60.85 & 59.19 \\
 & Seed 456 & 99.50 & 99.50 & 94.19 & 94.18 & 70.57 & 70.35 & 64.56 & 63.97 \\
\cmidrule(lr){2-10}
 & Mean & 99.50 & 99.50 & 94.89 & 94.89 & 70.20 & 69.73 & 63.48 & 62.54 \\
 & Std & 0.07 & 0.07 & 0.64 & 0.66 & 1.31 & 1.64 & 2.29 & 2.91 \\
\midrule
\multirow{5}{*}{DA-MTL} & Seed 42 & 99.59 & 99.59 & 95.54 & 95.54 & 71.73 & 71.25 & 65.83 & 64.46 \\
 & Seed 123 & 99.59 & 99.59 & 95.02 & 95.02 & 70.84 & 70.32 & 65.48 & 64.28 \\
 & Seed 456 & 99.56 & 99.56 & 95.37 & 95.37 & 70.68 & 70.75 & 63.66 & 63.32 \\
\cmidrule(lr){2-10}
 & Mean & 99.58 & 99.58 & 95.31 & 95.31 & 71.08 & 70.78 & 64.99 & 64.02 \\
 & Std & 0.02 & 0.02 & 0.27 & 0.26 & 0.56 & 0.47 & 1.16 & 0.61 \\
\midrule
\multirow{5}{*}{Idiosyncrasies} & Seed 42 & 99.49 & 99.49 & 94.42 & 94.44 & 71.06 & 70.63 & 65.65 & 64.70 \\
 & Seed 123 & 99.69 & 99.69 & 94.31 & 94.33 & 72.28 & 72.07 & 64.90 & 64.05 \\
 & Seed 456 & 99.51 & 99.51 & 96.39 & 96.38 & 72.60 & 72.13 & 65.26 & 63.59 \\
\cmidrule(lr){2-10}
 & Mean & 99.56 & 99.56 & 95.04 & 95.05 & 71.98 & 71.61 & 65.27 & 64.11 \\
 & Std & 0.11 & 0.11 & 1.17 & 1.16 & 0.81 & 0.85 & 0.37 & 0.56 \\
\midrule
\multirow{5}{*}{\mname{} (ours)} & Seed 42 & 99.77 & 99.77 & 98.11 & 98.12 & 84.36 & 84.33 & 74.30 & 73.96 \\
 & Seed 123 & 99.72 & 99.72 & 98.01 & 98.01 & 84.20 & 84.20 & 74.50 & 74.31 \\
 & Seed 456 & 99.68 & 99.68 & 97.78 & 97.78 & 84.33 & 84.29 & 74.00 & 73.70 \\
\cmidrule(lr){2-10}
 & Mean & 99.72 & 99.72 & 97.97 & 97.97 & 84.30 & 84.27 & 74.27 & 73.99 \\
 & Std & 0.05 & 0.05 & 0.17 & 0.17 & 0.08 & 0.07 & 0.25 & 0.31 \\
\bottomrule
\end{tabular}
\end{table}

\clearpage
\subsection{OpenTuringBench Setup}
\label{app:exp_otb}

OpenTuringBench~\citep{openturingbench} consists of news articles generated by
seven open language models, along with human-written articles, and provides
$264{,}321$ training texts in total.
For our MAA
setting, we use the machine-generated texts and treat the seven source models as the candidate classes, excluding the human-written class.
Inputs are truncated to
$1024$ tokens, compared with $512$ tokens for \benchname{}.
The seven source models range from $3.8$B to $10.7$B parameters:
{Phi-3-mini-4k-instruct}, {Neural-chat-7b-v3-3},
{Mistral-7B-Instruct-v0.3}, {Qwen2.5-7B-Instruct},
{Llama-3.1-8B-Instruct}, {gemma-2-9b-it}, and
{SOLAR-10.7B-Instruct-v1.0}.
Table~\ref{tab:otb_tasks} lists the evaluation tasks.

\begin{table}[ht]
\centering
\caption{\textbf{OpenTuringBench tasks.}}
\label{tab:otb_tasks}
\footnotesize
\setlength{\tabcolsep}{4pt}
\begin{tabular}{l l p{8.2cm}}
\toprule
\textbf{Reported as} & \textbf{Task} & \textbf{What changes} \\
\midrule
ID  & E0 & Held-out news articles from the training source \\
E1  & E1 & Sampling temperature raised to $0.7$ and $1.0$ \\
E2  & E2 & $\sim$70B versions of Llama 3.1 and Qwen 2.5 in place of the 8B and 7B ones seen in training \\
E3  & E3 & Each model rewrites its own earlier generations \\
E4  & E4 & Human--machine mixing: models revise and continue human-written text \\
OOD & E5 & Essay domain instead of news, about $1{,}000$ essays per model \\
\bottomrule
\end{tabular}
\end{table}

We here provide Table~\ref{tab:otb_f1}, which reports macro-F1 for the same runs done in Table~\ref{tab:otb_results}, which reports accuracy only.
It also implies that \mname{} still outperforms all baselines on every task except ID.

\begin{table}[!ht]
\centering
\footnotesize
\setlength{\tabcolsep}{4.2pt}
\renewcommand{\arraystretch}{1.05}

\caption{
\textbf{MAA performance on OpenTuringBench (macro-F1).}
We report macro-F1 ($\uparrow$) for the runs of Table~\ref{tab:otb_results}.
Best results are in \textbf{bold}, and second-best results are \second{underlined}.
}
\label{tab:otb_f1}

\begin{tabular}{
@{}l
@{\hspace{8pt}}
ccccc
@{\hspace{8pt}}
c@{}
}
\toprule

\multirow{2}{*}{Method}
& \multicolumn{5}{c}{ID}
& \multirow{2}{*}{OOD} \\

\cmidrule(lr){2-6}

& ID & E1 & E2 & E3 & E4 & \\

\midrule

Fast-DetectGPT~\citep{fastdetectgpt}
& 14.06 & 16.79 & 13.23 & 17.13 & 7.76
& 13.91 \\

Binoculars~\citep{binoculars}
& 13.41 & 18.28 & 13.88 & 20.88 & 9.26
& 19.42 \\

GPT-who~\citep{gpt-who}
& 47.00 & 16.49 & 43.80 & 29.82 & 6.77
& 24.25 \\

Sniffer~\citep{sniffer}
& 61.48 & 18.15 & 54.94 & 37.42 & 7.28
& 27.68 \\

StatEnsemble~\citep{multilingual}
& 61.18 & 60.58 & 57.42 & 63.06 & 26.13
& 48.26 \\

\addlinespace[2pt]

DeTeCtive~\citep{detective}
& \second{99.83} & \second{83.81} & 99.36 & 88.93 & 20.87
& 52.82 \\

OTBDetector~\citep{openturingbench}
& 99.76 & 83.53 & 99.11 & 88.39 & 25.06
& 45.12 \\

DA-MTL~\citep{damtl}
& \best{99.87} & 82.90 & \second{99.48} & \second{96.02} & \second{27.06}
& \second{58.46} \\

Idiosyncrasies~\citep{idiosyncrasies}
& 88.33 & 67.00 & 85.04 & 77.90 & 23.86
& 35.97 \\

\midrule
\rowcolor{gray!10}
\mname{} (ours)
& 99.79 & \best{89.51} & \best{99.70} & \best{98.78} & \best{41.88}
& \best{95.58} \\

\bottomrule
\end{tabular}
\end{table}

\clearpage
\subsection{MGT Detection}
\label{app:exp_mgt}

We here report the additional metrics considered in the MGT detection task.
That is, for the same runs of Table~\ref{tab:mgt_results} (which reports AUROC), Table~\ref{tab:mgt_tpr} shows the true positive rate at a $1\%$ false positive rate (TPR@1\%FPR) and the false positive rate at a $95\%$ true positive rate (FPR@95\%TPR).
The results also suggest that \mname{} can be applied well to the MGT detection task.

\begin{table*}[!ht]
\centering
\footnotesize
\setlength{\tabcolsep}{3.0pt}
\renewcommand{\arraystretch}{1.05}
\caption{
\textbf{MGT detection on \benchname{} at fixed operating points.}
For the runs of Table~\ref{tab:mgt_results}, we report the true positive rate at a $1\%$ false positive rate (TPR, $\uparrow$) and the false positive rate at a $95\%$ true positive rate (FPR, $\downarrow$), with MGT as the positive class; \emph{Trained} marks methods trained on \benchname{}, as in Table~\ref{tab:mgt_results}.
Best results are in \textbf{bold}.
}
\label{tab:mgt_tpr}

\begin{tabular}{
@{}l
@{\hspace{5pt}}
c
@{\hspace{5pt}}
cc
@{\hspace{5pt}}
cc
@{\hspace{5pt}}
cc
@{\hspace{5pt}}
cc@{}
}
\toprule

\multirow{3}{*}{Method}
& \multirow{3}{*}{Trained}
& \multicolumn{2}{c}{\multirow{2}{*}{ID}}
& \multicolumn{6}{c}{OOD} \\
\cmidrule(lr){5-10}

& &
& & \multicolumn{2}{c}{Domain}
& \multicolumn{2}{c}{Version}
& \multicolumn{2}{c}{Joint} \\

\cmidrule(lr){3-4}
\cmidrule(lr){5-6}
\cmidrule(lr){7-8}
\cmidrule(lr){9-10}

& & TPR & FPR & TPR & FPR & TPR & FPR & TPR & FPR \\

\midrule

Fast-DetectGPT~\citep{fastdetectgpt}
& $\times$
& 17.31 & 95.04 & 24.90 & 95.78 & 13.14 & 96.31 & 13.35 & 98.82 \\

Binoculars~\citep{binoculars}
& $\times$
& 26.13 & 91.29 & 18.31 & 83.94 & 19.75 & 93.99 & 11.76 & 93.48 \\

IRM~\citep{irm}
& $\times$
& 18.09 & 78.27 & 15.39 & 90.09 & 13.25 & 80.35 & 10.41 & 92.85 \\

ReMoDetect~\citep{remodetect}
& $\times$
& 17.65 & 78.97 & 31.41 & 71.95 & 15.11 & 82.17 & 29.72 & 75.09 \\

\addlinespace[2pt]
DeTeCtive~\citep{detective}
& \checkmark
& 99.98 & \best{0.00} & 96.51 & 0.13 & 99.77 & \best{0.00} & 93.52 & 7.29 \\

OTBDetector~\citep{openturingbench}
& \checkmark
& 99.98 & 0.10 & 98.90 & 0.07 & 99.57 & 0.10 & 94.29 & 10.41 \\

DA-MTL~\citep{damtl}
& \checkmark
& \best{100.00} & \best{0.00} & 98.78 & \best{0.00} & \best{99.97} & \best{0.00} & \best{98.13} & \best{0.00} \\

Idiosyncrasies~\citep{idiosyncrasies}
& \checkmark
& \best{100.00} & \best{0.00} & 98.55 & \best{0.00} & \best{99.97} & \best{0.00} & 95.79 & 0.28 \\

\midrule
\rowcolor{gray!10}
\mname{} (ours)
& \checkmark
& \best{100.00} & \best{0.00} & \best{99.45} & \best{0.00} & 99.93 & \best{0.00} & 97.25 & 0.21 \\

\bottomrule
\end{tabular}
\end{table*}

\clearpage
\subsection{Ablation and Scaling Analysis}
\label{app:exp_analysis}

\paragraph{Source modeling and likelihood aggregation.}

We train each method for 10,000 texts per source for five epochs on the 7B backbone.
Table~\ref{tab:components_full} shows the results, suggesting that both components mainly help when the source-model version changes.
On ID and Domain shift, the three variants stay within $0.7$pp of each other, so neither component is needed to fit the training distribution or to transfer across domains.
On Version shift and Joint shift, source-specific up-projections improve accuracy by $2.18$pp and $0.84$pp over a shared $B$, and clipped scoring adds a further $4.55$pp and $5.12$pp, for a total gain of $6.73$pp and $5.96$pp over the shared-$B$, mean-score variant.
Clipping therefore accounts for most of the gain, consistent with its role of preventing a few tokens with extreme likelihood differences from deciding the attribution when the source model changes.

\begin{table}[ht]
\centering
\footnotesize
\setlength{\tabcolsep}{4.2pt}
\renewcommand{\arraystretch}{1.05}
\caption{
\textbf{Component analysis on all four evaluation settings of \benchname{}.}
We report accuracy ($\uparrow$) with $10{,}000$ training texts per source family. \emph{Up-projection} is shared across sources or per source, and \emph{Scoring} is the mean likelihood or the clipped score.
Best results are in \textbf{bold}.
}
\label{tab:components_full}

\begin{tabular}{
@{}l
@{\hspace{8pt}}
l
@{\hspace{8pt}}
c
@{\hspace{5pt}}
ccc@{}
}
\toprule

\multirow{2}{*}{Up-projection}
& \multirow{2}{*}{Scoring}
& \multirow{2}{*}{ID}
& \multicolumn{3}{c}{OOD} \\

\cmidrule(lr){4-6}

& & & Domain & Version & Joint \\

\midrule

shared $B$       & mean    & 99.70 & 97.51 & 76.06 & 66.11 \\
per-source $B_k$ & mean    & \best{99.86} & 96.91 & 78.24 & 66.95 \\

\midrule
\rowcolor{gray!10}
per-source $B_k$ & clipped & 99.64 & \best{97.59} & \best{82.79} & \best{72.07} \\

\bottomrule
\end{tabular}
\end{table}

\paragraph{Effect of backbone model size.}

Table~\ref{tab:backbone_scale} gives the exact accuracy and macro-F1 behind Figure~\ref{fig:backbone_scale} (a), which shows similar trends of performance improvement with respect to the backbone size.
Each backbone is trained on 10,000 texts per source family for five epochs.

\begin{table*}[!ht]
\centering
\footnotesize
\setlength{\tabcolsep}{4.2pt}
\renewcommand{\arraystretch}{1.05}
\caption{
\textbf{Effect of backbone size.}
We report accuracy (Acc, $\uparrow$) and macro-F1 (F1, $\uparrow$) for Qwen2.5 backbones of four sizes, each trained on $10{,}000$ texts per source family; the last row is the change in accuracy from 0.5B to 7B.
Best results are in \textbf{bold}.
}
\label{tab:backbone_scale}

\begin{tabular}{
@{}l
@{\hspace{8pt}}
cc
@{\hspace{5pt}}
cc
@{\hspace{5pt}}
cc
@{\hspace{5pt}}
cc@{}
}
\toprule

\multirow{3}{*}{Backbone}
& \multicolumn{2}{c}{\multirow{2}{*}{ID}}
& \multicolumn{6}{c}{OOD} \\
\cmidrule(lr){4-9}

&
&
& \multicolumn{2}{c}{Domain}
& \multicolumn{2}{c}{Version}
& \multicolumn{2}{c}{Joint} \\

\cmidrule(lr){2-3}
\cmidrule(lr){4-5}
\cmidrule(lr){6-7}
\cmidrule(lr){8-9}

& Acc & F1
& Acc & F1
& Acc & F1
& Acc & F1 \\

\midrule

Qwen2.5-0.5B
& 98.32 & 98.32
& 88.19 & 88.28
& 78.33 & 78.54
& 64.98 & 64.92 \\

Qwen2.5-1.5B
& 99.01 & 99.01
& 96.17 & 96.18
& 81.00 & 81.08
& 70.28 & 70.21 \\

Qwen2.5-3B
& 99.38 & 99.38
& 94.59 & 94.63
& 82.18 & 82.20
& 71.21 & 71.33 \\

Qwen2.5-7B
& \best{99.64} & \best{99.64}
& \best{97.59} & \best{97.60}
& \best{82.79} & \best{82.75}
& \best{72.07} & \best{71.78} \\

\midrule

$\Delta$ 7B $-$ 0.5B
& \multicolumn{2}{c}{+1.32pp}
& \multicolumn{2}{c}{+9.40pp}
& \multicolumn{2}{c}{+4.46pp}
& \multicolumn{2}{c}{+7.09pp} \\

\bottomrule
\end{tabular}
\end{table*}

Further, Table~\ref{tab:backbone_family} replaces the backbone with OLMo-2-7B and Qwen3-8B under the same training setup of 10,000 texts per source.
The three backbones stay within about 1pp of each other on ID, Domain shift, and Version shift, and Qwen3-8B improves Joint shift to 75.22\%, so we can conclude that the gains of \mname{} do not depend on a particular backbone.

\begin{table*}[!ht]
\centering
\footnotesize
\setlength{\tabcolsep}{4.2pt}
\renewcommand{\arraystretch}{1.05}
\caption{
\textbf{Effect of backbone family.}
We report accuracy (Acc, $\uparrow$) and macro-F1 (F1, $\uparrow$) for three backbones of comparable size, each trained on $10{,}000$ texts per source family.
Best results are in \textbf{bold}.
}
\label{tab:backbone_family}

\begin{tabular}{
@{}l
@{\hspace{8pt}}
cc
@{\hspace{5pt}}
cc
@{\hspace{5pt}}
cc
@{\hspace{5pt}}
cc@{}
}
\toprule

\multirow{3}{*}{Backbone}
& \multicolumn{2}{c}{\multirow{2}{*}{ID}}
& \multicolumn{6}{c}{OOD} \\
\cmidrule(lr){4-9}

&
&
& \multicolumn{2}{c}{Domain}
& \multicolumn{2}{c}{Version}
& \multicolumn{2}{c}{Joint} \\

\cmidrule(lr){2-3}
\cmidrule(lr){4-5}
\cmidrule(lr){6-7}
\cmidrule(lr){8-9}

& Acc & F1
& Acc & F1
& Acc & F1
& Acc & F1 \\

\midrule

Qwen2.5-7B
& 99.64 & 99.64
& 97.59 & 97.60
& 82.79 & 82.75
& 72.07 & 71.78 \\

OLMo-2-7B
& 99.60 & 99.60
& 97.05 & 97.07
& \best{83.75} & \best{83.79}
& 72.99 & 73.07 \\

Qwen3-8B
& \best{99.73} & \best{99.73}
& \best{98.09} & \best{98.09}
& 83.40 & 83.26
& \best{75.22} & \best{74.88} \\

\bottomrule
\end{tabular}
\end{table*}

\clearpage
\paragraph{Effect of training data size.}

Table~\ref{tab:data_scale} shows that accuracy improves with training size on all four evaluation settings, and more so under shift: from 1k texts per source to the full split, ID accuracy rises by 1.12pp against 7.10pp on Joint shift.
Only Domain shift is non-monotonic, dipping at 5k before recovering.
Additional training data therefore buys robustness rather than ID accuracy, the same pattern that the architectural components and the backbone scale show.

\begin{table*}[!ht]
\centering
\footnotesize
\setlength{\tabcolsep}{4.2pt}
\renewcommand{\arraystretch}{1.05}
\caption{
\textbf{Effect of training-data scale.}
We report accuracy ($\uparrow$) against the number of training texts per source family; the \emph{full} column of every method is its entry in Table~\ref{tab:main_results}.
Best results in each column of a block are in \textbf{bold}.
}
\label{tab:data_scale}

\begin{tabular}{
@{}l
@{\hspace{8pt}}
cccccc@{}
}
\toprule

\multirow{2}{*}{Method}
& \multicolumn{6}{c}{Training texts per source} \\
\cmidrule(lr){2-7}
& 1k & 3k & 5k & 10k & 15k & full \\

\midrule

\multicolumn{7}{@{}l}{\textit{ID}} \\
DeTeCtive~\citep{detective} & 96.25 & 98.08 & 98.52 & 99.20 & 99.28 & 99.49 \\
OTBDetector~\citep{openturingbench} & 48.64 & 73.83 & 81.87 & 98.29 & 99.07 & 99.42 \\
DA-MTL~\citep{damtl} & 95.91 & 98.34 & 98.92 & 99.30 & 99.41 & 99.45 \\
Idiosyncrasies~\citep{idiosyncrasies} & 95.44 & 98.41 & 98.90 & 99.03 & 99.30 & 99.52 \\
\rowcolor{gray!10}
\mname{} (ours) & \best{98.65} & \best{99.41} & \best{99.55} & \best{99.64} & \best{99.73} & \best{99.77} \\

\addlinespace[2pt]

\multicolumn{7}{@{}l}{\textit{Domain Shift}} \\
DeTeCtive~\citep{detective} & 83.87 & 88.67 & 94.08 & 95.21 & 93.48 & 95.38 \\
OTBDetector~\citep{openturingbench} & 35.61 & 63.71 & 82.10 & 91.62 & 92.44 & 95.44 \\
DA-MTL~\citep{damtl} & 84.08 & 91.86 & 94.94 & 94.02 & 94.12 & 94.59 \\
Idiosyncrasies~\citep{idiosyncrasies} & 81.98 & 92.34 & 93.61 & 92.41 & 93.61 & 94.39 \\
\rowcolor{gray!10}
\mname{} (ours) & \best{96.36} & \best{97.32} & \best{96.89} & \best{97.59} & \best{97.74} & \best{98.11} \\

\addlinespace[2pt]

\multicolumn{7}{@{}l}{\textit{Version Shift}} \\
DeTeCtive~\citep{detective} & 60.10 & 64.89 & 65.44 & 67.05 & 67.74 & 70.08 \\
OTBDetector~\citep{openturingbench} & 31.67 & 49.02 & 60.15 & 68.77 & 71.41 & 71.27 \\
DA-MTL~\citep{damtl} & 59.82 & 66.87 & 66.72 & 67.43 & 69.38 & 72.78 \\
Idiosyncrasies~\citep{idiosyncrasies} & 60.04 & 64.91 & 65.98 & 66.03 & 68.86 & 71.02 \\
\rowcolor{gray!10}
\mname{} (ours) & \best{80.10} & \best{81.05} & \best{81.95} & \best{82.79} & \best{83.42} & \best{84.36} \\

\addlinespace[2pt]

\multicolumn{7}{@{}l}{\textit{Joint Shift}} \\
DeTeCtive~\citep{detective} & 56.95 & 57.35 & 61.81 & 62.51 & 62.13 & 62.92 \\
OTBDetector~\citep{openturingbench} & 33.39 & 51.10 & 58.24 & 62.95 & 64.24 & 65.02 \\
DA-MTL~\citep{damtl} & 55.96 & 62.32 & 63.46 & 63.81 & 64.31 & 66.15 \\
Idiosyncrasies~\citep{idiosyncrasies} & 56.74 & 60.94 & 60.43 & 60.15 & 61.42 & 65.66 \\
\rowcolor{gray!10}
\mname{} (ours) & \best{67.20} & \best{69.01} & \best{69.32} & \best{72.07} & \best{73.08} & \best{74.30} \\

\bottomrule
\end{tabular}
\end{table*}

\clearpage
\subsection{Additional Analysis}
\label{app:exp_additional}

Both analyses in this section use the seed-$42$ run of each method (Table~\ref{tab:main_seeds}).

\paragraph{Confusion matrices.}
Figure~\ref{fig:confusion} breaks the accuracy of \mname{} down by source family.
On ID and Domain shift, every family keeps a recall above $97\%$.
Under Version shift and Joint shift, the errors concentrate on two families and mostly go to Claude: $32.5\%$ and $52.5\%$ of Gemini texts and $13.7\%$ and $24.2\%$ of GPT texts are attributed to Claude, whereas Claude itself keeps a recall above $92\%$.
Mistral and Llama are less affected, with recalls of at least $76.2\%$.

\newcommand{\cmcell}[3]{%
  \fill[bbNavy!#3] (#1*0.92,-#2*0.5) rectangle (#1*0.92+0.90,-#2*0.5-0.48);
  \ifdim#3pt>50pt
    \node[font=\tiny,text=white] at (#1*0.92+0.45,-#2*0.5-0.24) {#3};
  \else
    \node[font=\tiny,text=black!80] at (#1*0.92+0.45,-#2*0.5-0.24) {#3};
  \fi}
\newcommand{\cmframe}[1]{%
  \node[anchor=south,font=\footnotesize\bfseries] at (2.29,0.62) {#1};
  \node[anchor=south,font=\scriptsize,text=black!60] at (2.29,0.30) {Predicted};
  \node[anchor=south,font=\scriptsize,text=black!60,rotate=90] at (-1.15,-1.24) {True};
  \foreach \n [count=\i from 0] in {GPT,Claude,Gemini,Mistral,Llama}{
    \node[anchor=east,font=\scriptsize] at (-0.06,-\i*0.5-0.24) {\n};
    \node[anchor=south,font=\scriptsize,rotate=0] at (\i*0.92+0.45,0.02) {\n};}}
\begin{figure}[H]
\centering
\begin{minipage}[t]{0.49\linewidth}\centering
{\sffamily\begin{tikzpicture}
\cmframe{ID}
\cmcell{0}{0}{100.0}
\cmcell{1}{0}{0.0}
\cmcell{2}{0}{0.0}
\cmcell{3}{0}{0.0}
\cmcell{4}{0}{0.0}
\cmcell{0}{1}{0.0}
\cmcell{1}{1}{99.8}
\cmcell{2}{1}{0.2}
\cmcell{3}{1}{0.0}
\cmcell{4}{1}{0.0}
\cmcell{0}{2}{0.0}
\cmcell{1}{2}{0.0}
\cmcell{2}{2}{99.6}
\cmcell{3}{2}{0.3}
\cmcell{4}{2}{0.1}
\cmcell{0}{3}{0.0}
\cmcell{1}{3}{0.0}
\cmcell{2}{3}{0.1}
\cmcell{3}{3}{99.4}
\cmcell{4}{3}{0.5}
\cmcell{0}{4}{0.0}
\cmcell{1}{4}{0.0}
\cmcell{2}{4}{0.0}
\cmcell{3}{4}{0.1}
\cmcell{4}{4}{99.9}
\end{tikzpicture}}
\end{minipage}\hfill
\begin{minipage}[t]{0.49\linewidth}\centering
{\sffamily\begin{tikzpicture}
\cmframe{Domain Shift}
\cmcell{0}{0}{97.4}
\cmcell{1}{0}{0.7}
\cmcell{2}{0}{1.9}
\cmcell{3}{0}{0.0}
\cmcell{4}{0}{0.0}
\cmcell{0}{1}{0.0}
\cmcell{1}{1}{98.2}
\cmcell{2}{1}{1.7}
\cmcell{3}{1}{0.1}
\cmcell{4}{1}{0.0}
\cmcell{0}{2}{0.0}
\cmcell{1}{2}{0.0}
\cmcell{2}{2}{99.4}
\cmcell{3}{2}{0.3}
\cmcell{4}{2}{0.3}
\cmcell{0}{3}{0.0}
\cmcell{1}{3}{0.0}
\cmcell{2}{3}{1.1}
\cmcell{3}{3}{97.3}
\cmcell{4}{3}{1.5}
\cmcell{0}{4}{0.0}
\cmcell{1}{4}{0.0}
\cmcell{2}{4}{0.3}
\cmcell{3}{4}{1.4}
\cmcell{4}{4}{98.3}
\end{tikzpicture}}
\end{minipage}
\vspace{8pt}

\begin{minipage}[t]{0.49\linewidth}\centering
{\sffamily\begin{tikzpicture}
\cmframe{Version Shift}
\cmcell{0}{0}{85.0}
\cmcell{1}{0}{13.7}
\cmcell{2}{0}{0.2}
\cmcell{3}{0}{1.2}
\cmcell{4}{0}{0.0}
\cmcell{0}{1}{7.5}
\cmcell{1}{1}{92.1}
\cmcell{2}{1}{0.2}
\cmcell{3}{1}{0.1}
\cmcell{4}{1}{0.1}
\cmcell{0}{2}{0.1}
\cmcell{1}{2}{32.5}
\cmcell{2}{2}{61.6}
\cmcell{3}{2}{5.2}
\cmcell{4}{2}{0.7}
\cmcell{0}{3}{1.5}
\cmcell{1}{3}{1.3}
\cmcell{2}{3}{7.5}
\cmcell{3}{3}{87.0}
\cmcell{4}{3}{2.8}
\cmcell{0}{4}{0.0}
\cmcell{1}{4}{0.0}
\cmcell{2}{4}{0.1}
\cmcell{3}{4}{3.8}
\cmcell{4}{4}{96.1}
\end{tikzpicture}}
\end{minipage}\hfill
\begin{minipage}[t]{0.49\linewidth}\centering
{\sffamily\begin{tikzpicture}
\cmframe{Joint Shift}
\cmcell{0}{0}{71.8}
\cmcell{1}{0}{24.2}
\cmcell{2}{0}{2.5}
\cmcell{3}{0}{1.2}
\cmcell{4}{0}{0.3}
\cmcell{0}{1}{4.4}
\cmcell{1}{1}{92.8}
\cmcell{2}{1}{2.4}
\cmcell{3}{1}{0.3}
\cmcell{4}{1}{0.2}
\cmcell{0}{2}{0.2}
\cmcell{1}{2}{52.5}
\cmcell{2}{2}{41.0}
\cmcell{3}{2}{5.6}
\cmcell{4}{2}{0.7}
\cmcell{0}{3}{5.3}
\cmcell{1}{3}{3.1}
\cmcell{2}{3}{9.3}
\cmcell{3}{3}{76.2}
\cmcell{4}{3}{6.1}
\cmcell{0}{4}{0.0}
\cmcell{1}{4}{0.0}
\cmcell{2}{4}{0.7}
\cmcell{3}{4}{9.6}
\cmcell{4}{4}{89.8}
\end{tikzpicture}}
\end{minipage}
\caption{\textbf{Confusion matrices of \mname{} on \benchname{}.} Each row is a true source family and each column a predicted one; entries are the percentage of the row's texts, so the diagonal is the per-family recall.}
\label{fig:confusion}
\end{figure}

\paragraph{Accuracy by model version.}
Figure~\ref{fig:version_order} breaks Version shift accuracy down by held-out version, ordered from the oldest to the newest release in each family, and compares \mname{} with OTBDetector.
Both methods lose accuracy as the version moves further from the one used for training, but \mname{} stays ahead on every version except the newest GPT version, where the two are within $1.5$pp.

\definecolor{famS}{HTML}{9C8AB8}\definecolor{famL}{HTML}{7FA88A}
\pgfplotsset{vo/.style={
  width=2.15cm, height=3.0cm, scale only axis,
  xmin=0.7, xmax=3.3, xtick={1,2,3}, xticklabels={1st,2nd,3rd},
  ymin=0, ymax=100, ytick={0,20,40,60,80,100},
  axis lines*=left, xmajorgrids=true, ymajorgrids=true,
  grid style={black!22,line width=0.4pt},
  axis line style={black!55}, tick style={black!55}, tick align=outside,
  tick label style={font=\scriptsize}, label style={font=\footnotesize},
  title style={font=\footnotesize\bfseries, yshift=-4pt},
  mark size=1.6pt, line width=1.0pt, clip=false,
}}
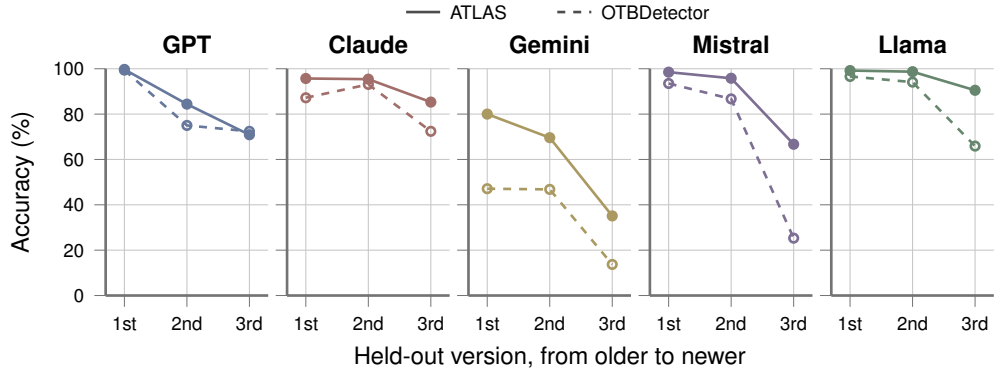
\begin{figure}[H]
\centering
{\sffamily
\begin{tikzpicture}
\begin{axis}[vo, name=p0,
  title={GPT}, ylabel={Accuracy (\%)}, yticklabels={0,20,40,60,80,100},
]
\addplot[famG!80!black,mark=*] coordinates {(1,99.7)(2,84.4)(3,70.9)};
\addplot[famG!80!black,dashed,mark=o,mark options={solid}] coordinates {(1,99.4)(2,75.0)(3,72.4)};
\end{axis}
\begin{axis}[vo, name=p1, at={(p0.south east)}, anchor=south west, xshift=0.25cm,
  title={Claude}, yticklabels={},
]
\addplot[famC!80!black,mark=*] coordinates {(1,95.7)(2,95.4)(3,85.3)};
\addplot[famC!80!black,dashed,mark=o,mark options={solid}] coordinates {(1,87.2)(2,93.1)(3,72.4)};
\end{axis}
\begin{axis}[vo, name=p2, at={(p1.south east)}, anchor=south west, xshift=0.25cm,
  title={Gemini}, yticklabels={},
]
\addplot[famM!80!black,mark=*] coordinates {(1,80.0)(2,69.6)(3,35.1)};
\addplot[famM!80!black,dashed,mark=o,mark options={solid}] coordinates {(1,47.1)(2,46.8)(3,13.7)};
\end{axis}
\begin{axis}[vo, name=p3, at={(p2.south east)}, anchor=south west, xshift=0.25cm,
  title={Mistral}, yticklabels={},
]
\addplot[famS!80!black,mark=*] coordinates {(1,98.5)(2,95.8)(3,66.7)};
\addplot[famS!80!black,dashed,mark=o,mark options={solid}] coordinates {(1,93.5)(2,86.7)(3,25.3)};
\end{axis}
\begin{axis}[vo, name=p4, at={(p3.south east)}, anchor=south west, xshift=0.25cm,
  title={Llama}, yticklabels={},
]
\addplot[famL!80!black,mark=*] coordinates {(1,99.2)(2,98.7)(3,90.5)};
\addplot[famL!80!black,dashed,mark=o,mark options={solid}] coordinates {(1,96.6)(2,94.1)(3,65.9)};
\end{axis}
\node[anchor=north,font=\footnotesize] at ([yshift=-0.55cm]p2.south) {Held-out version, from older to newer};
\draw[black!70,line width=1.0pt] ([xshift=-1.9cm,yshift=0.75cm]p2.north) -- ++(0.45,0) node[right,font=\scriptsize,text=black] {\mname{}};
\draw[black!70,line width=1.0pt,dashed] ([xshift=0.1cm,yshift=0.75cm]p2.north) -- ++(0.45,0) node[right,font=\scriptsize,text=black] {OTBDetector};
\end{tikzpicture}}
\caption{\textbf{Version Shift accuracy by held-out version.} Accuracy ($\uparrow$) of \mname{} (solid) and OTBDetector (dashed) on each of the three held-out versions of a family, ordered from the oldest to the newest release; Table~\ref{tab:bench_source_models} lists the versions.}
\label{fig:version_order}
\end{figure}

\clearpage
\raggedbottom
\section{Beyond Attribution: Distilled and Deployed Models}
\label{app:exp_distillation}

\paragraph{Attribution of distilled and deployed models.}

\mname{} is trained as a three-class attributor over GPT, Claude, and Gemini for five epochs under the settings of Appendix~\ref{app:exp_training}.
Its standardization uses the $3{,}994$ Mistral and Llama texts of the $1{,}997$ ID test seeds of \benchname{}, none of which is used in training.
The distilled models each generate one text per ID test seed with the same prompt template, served locally at temperature $0.7$ and at most $1{,}024$ new tokens; where a model emits a reasoning trace, only the final answer is scored.
The seven deployed models of Figure~\ref{fig:distill} are queried through the DeepInfra API and Grok-4.7~\citep{grok} through the LiteLLM API, while A.X-K2~\citep{axk2}, K-EXAONE-2.0~\citep{exaone}, Motif-3~\citep{motif3}, and Solar-Open2~\citep{solar} are served locally from their public checkpoints (\texttt{skt/A.X-K2}, \texttt{LGAI-EXAONE/K-EXAONE-2.0-750B-A37B-FP8}, \texttt{Motif-Technologies/Motif-3}, and \texttt{upstage/Solar-Open2-250B}).
Table~\ref{tab:distill} reports the shares and reject rates of all models, including the comparison models, and Table~\ref{tab:distill_validation} lists the six distilled checkpoints.
\begin{table}[H]
\centering
\footnotesize
\setlength{\tabcolsep}{4.2pt}
\renewcommand{\arraystretch}{1.05}

\caption{
\textbf{Attribution of distilled and deployed models.}
Share of each model's texts attributed to each candidate family, in percent.
Each validation model is listed by the base model it was fine-tuned from, with the family of the teacher stated, and the most attributed family is in \textbf{bold}.
\emph{Reject} is the share of texts that no condition explains better than the frozen backbone.
}
\label{tab:distill}

\begin{tabular}{
@{}l
@{\hspace{8pt}}
ccc
@{\hspace{8pt}}
c@{}
}
\toprule

\multirow{2}{*}{Model}
& \multicolumn{3}{c}{Attribution (\%)}
& \multirow{2}{*}{Reject (\%)} \\

\cmidrule(lr){2-4}

& GPT & Claude & Gemini & \\

\midrule

\multicolumn{5}{@{}l}{\textit{Base model (teacher)}} \\
\phantom{000}Qwen3-8B (GPT)          & \best{93.6} & 5.3  & 1.1 & 10.6 \\
\phantom{000}Llama-3.1-8B (GPT)      & \best{87.3} & 11.9 & 0.8 & 30.2 \\
\phantom{000}Qwen3.5-27B (Claude)    & 5.8  & \best{92.9} & 1.3 & 1.8 \\
\phantom{000}Qwen3.5-4B (Claude)     & 23.0 & \best{76.4} & 0.6 & 8.0 \\
\phantom{000}Qwen3-30B-A3B (Gemini)  & 13.4 & 10.6 & \best{76.0} & 0.6 \\
\phantom{000}Qwen3-8B (Gemini)       & 30.6 & 6.1  & \best{63.3} & 1.5 \\

\addlinespace[2pt]

\multicolumn{5}{@{}l}{\textit{Deployed models (resemblance)}} \\
\phantom{000}DeepSeek-V4-Pro    & \best{90.6} & 9.3  & 0.1 & 0.1 \\
\phantom{000}DeepSeek-V3.2      & \best{41.5} & 27.8 & 30.6 & 0.4 \\
\phantom{000}GLM-5.2            & 15.2 & \best{84.7} & 0.1 & 0.0 \\
\phantom{000}GLM-5.1            & 31.5 & \best{63.6} & 4.9 & 0.2 \\
\phantom{000}MiniMax-M3         & \best{56.4} & 43.6 & 0.1 & 0.2 \\
\phantom{000}MiniMax-M2.7-Turbo & 6.2  & \best{93.8} & 0.0 & 0.1 \\
\phantom{000}Kimi-K2.6          & 14.8 & \best{85.2} & 0.0 & 0.1 \\
\phantom{000}Grok-4.7                & \best{92.2} & 7.8  & 0.1 & 0.2 \\
\phantom{000}A.X-K2                  & \best{62.6} & 37.2 & 0.2 & 0.2 \\
\phantom{000}K-EXAONE-2.0            & \best{72.6} & 26.1 & 1.4 & 1.5 \\
\phantom{000}Motif-3                 & \best{71.8} & 27.3 & 0.9 & 0.8 \\
\phantom{000}Solar-Open2             & \best{59.2} & 40.0 & 0.8 & 1.0 \\

\addlinespace[2pt]

\multicolumn{5}{@{}l}{\textit{Comparison models (no reported distillation from the three families)}} \\
\phantom{000}Falcon3-7B-Instruct     & \best{40.6} & 23.2 & 36.2 & 68.2 \\
\phantom{000}Granite-3.0-8B-Instruct & 37.7 & 18.7 & \best{43.7} & 74.6 \\

\bottomrule
\end{tabular}
\end{table}

\begin{table}[H]
\centering
\footnotesize
\setlength{\tabcolsep}{4.2pt}
\renewcommand{\arraystretch}{1.05}

\caption{
\textbf{Validation models for teacher identification.}
Publicly released checkpoints whose model cards state the teacher they were distilled from.
}
\label{tab:distill_validation}

\begin{minipage}{\linewidth}
\centering
\renewcommand{\thempfootnote}{\arabic{mpfootnote}}
\begin{tabular}{@{}l@{\hspace{5pt}}l@{}}
\toprule
Checkpoint & Stated teacher \\
\midrule
\texttt{TeichAI/Qwen3-8B-GPT-5.2-High-Reasoning-Distill}\footnote{\url{https://huggingface.co/TeichAI/Qwen3-8B-GPT-5.2-High-Reasoning-Distill}}              & GPT-5.2 \\
\texttt{Jackrong/GPT-5-Distill-llama3.1-8B-Instruct}\footnote{\url{https://huggingface.co/Jackrong/GPT-5-Distill-llama3.1-8B-Instruct}}                  & GPT-5 \\
\texttt{TeichAI/Qwen3.5-27B-Claude-Opus-4.6-Distill}\footnote{\url{https://huggingface.co/TeichAI/Qwen3.5-27B-Claude-Opus-4.6-Distill}}                  & Claude Opus 4.6 \\
\texttt{TeichAI/Qwen3.5-4B-Claude-Opus-Reasoning-Distill}\footnote{\url{https://huggingface.co/TeichAI/Qwen3.5-4B-Claude-Opus-Reasoning-Distill}}             & Claude Opus \\
\texttt{TeichAI/Qwen3-30B-A3B-Thinking-2507-Gemini-2.5-Flash-Distill}\footnote{\url{https://huggingface.co/TeichAI/Qwen3-30B-A3B-Thinking-2507-Gemini-2.5-Flash-Distill}} & Gemini 2.5 Flash \\
\texttt{TeichAI/Qwen3-8B-Gemini-2.5-Flash-Distill}\footnote{\url{https://huggingface.co/TeichAI/Qwen3-8B-Gemini-2.5-Flash-Distill}}                    & Gemini 2.5 Flash \\
\bottomrule
\end{tabular}
\end{minipage}
\end{table}

\end{document}